%% file: spie_manuscript.tex
\documentclass[]{spie}  

\input{tex/preamble}

\usepackage{amsmath,amsfonts,amssymb}
\usepackage{graphicx}
\usepackage[colorlinks=true, allcolors=blue]{hyperref}

\title{Self-Updating Models with Error Remediation}

\author[a]{Justin E. Doak (JD)}
\author[a]{Michael R. Smith (Mike)}
\author[a]{Joey B. Ingram (Joe)}

\affil[a]{Sandia National Laboratories, PO Box 5800, Albuquerque, NM USA}

\authorinfo{Further author information: J.E.D.: E-mail: jedoak@sandia.gov, Telephone: 1 505 844 1947}

\begin{document} 
\maketitle

\input{tex/abstract}

\keywords{self-updating models, label correction, autonomous model
  updating, autonomous machine learning, semi-supervised learning,
  model coupling, error propagation, feature-dependent label noise}

Justin E. Doak, Michael R. Smith, Joey B. Ingram, "Self-updating
models with error remediation," Proc. SPIE 11413, Artificial
Intelligence and Machine Learning for Multi-Domain Operations
Applications II, 114131W (18 May 2020); doi: 10.1117/12.2563843

Copyright 2020 Society of PhotoOptical Instrumentation Engineers. One
print or electronic copy may be made for personal use only. Systematic
reproduction and distribution, duplication of any material in this
paper for a fee or for commercial purposes, or modification of the
content of the paper are prohibited.

\section{INTRODUCTION}
\label{sec:intro}
\input{tex/introduction}

\section{PROBLEM FORMULATION}
\label{sec:problem}
\input{tex/problem}

\section{Experimental Methodology and Results}
\label{sec:experiments}
\input{tex/experiments}

\section{Related Work}
\label{sec:related}
\input{tex/related}

\section{Conclusion and Future Work}
\label{sec:conclusion}
\input{tex/conclusion}

\acknowledgments
\label{sec:acknowledgement}
\input{tex/acknowledgement}

\bibliography{spie_manuscript} 
\bibliographystyle{spiebib} 

\end{document}

%% file: tex/preamble.tex
\usepackage{amsmath}
\usepackage{amssymb}
\usepackage{array,graphicx}
\usepackage{booktabs}
\usepackage{pifont}
\usepackage[txcdraw,table]{xcolor}
\usepackage[symbol]{footmisc}
\usepackage{subfig}

\stepcounter{footnote}

\newcommand*\rot{\rotatebox{70}}
\newcommand*\OK{\ding{51}}
\definecolor{thunderbird}{RGB}{0, 172, 218}

\makeatletter
\let\old@mkpream\@mkpream
\def\@mkpream{%
\ifx\CT@drsc@\relax\else\let\CT@drsc@ @\fi
\let\CT@arc@\relax
\old@mkpream}
\makeatother

\usepackage{fancyhdr}

%% file: tex/abstract.tex
\begin{abstract}
Many environments currently employ machine learning models for data
processing and analytics that were built using a limited number of
training data points. Once deployed, the models are exposed to
significant amounts of previously-unseen data, not all of which is
representative of the original, limited training data. However,
updating these deployed models can be difficult due to logistical,
bandwidth, time, hardware, and/or data sensitivity constraints. We
propose a framework, Self-Updating Models with Error Remediation
(SUMER), in which a deployed model updates itself as new data becomes
available. SUMER uses techniques from semi-supervised learning and noise remediation to
iteratively retrain a deployed model using intelligently-chosen
predictions from the model as the labels for new training
iterations. A key component of SUMER is the notion of error
remediation as self-labeled data can be susceptible to the propagation
of errors. We investigate the use of SUMER across various data sets
and iterations. We find that self-updating models (SUMs) generally
perform better than models that do not attempt to self-update when
presented with additional previously-unseen data.  This performance
gap is accentuated in cases where there is only limited amounts of
initial training data. We also find that the performance of SUMER is
generally better than the performance of SUMs, demonstrating a benefit
in applying error remediation.  Consequently, SUMER can autonomously
enhance the operational capabilities of existing data processing
systems by intelligently updating models in dynamic environments.
\end{abstract}

%% file: tex/introduction.tex
\begin{figure}[!h]
  \centering
  \includegraphics[width=0.65\textwidth]{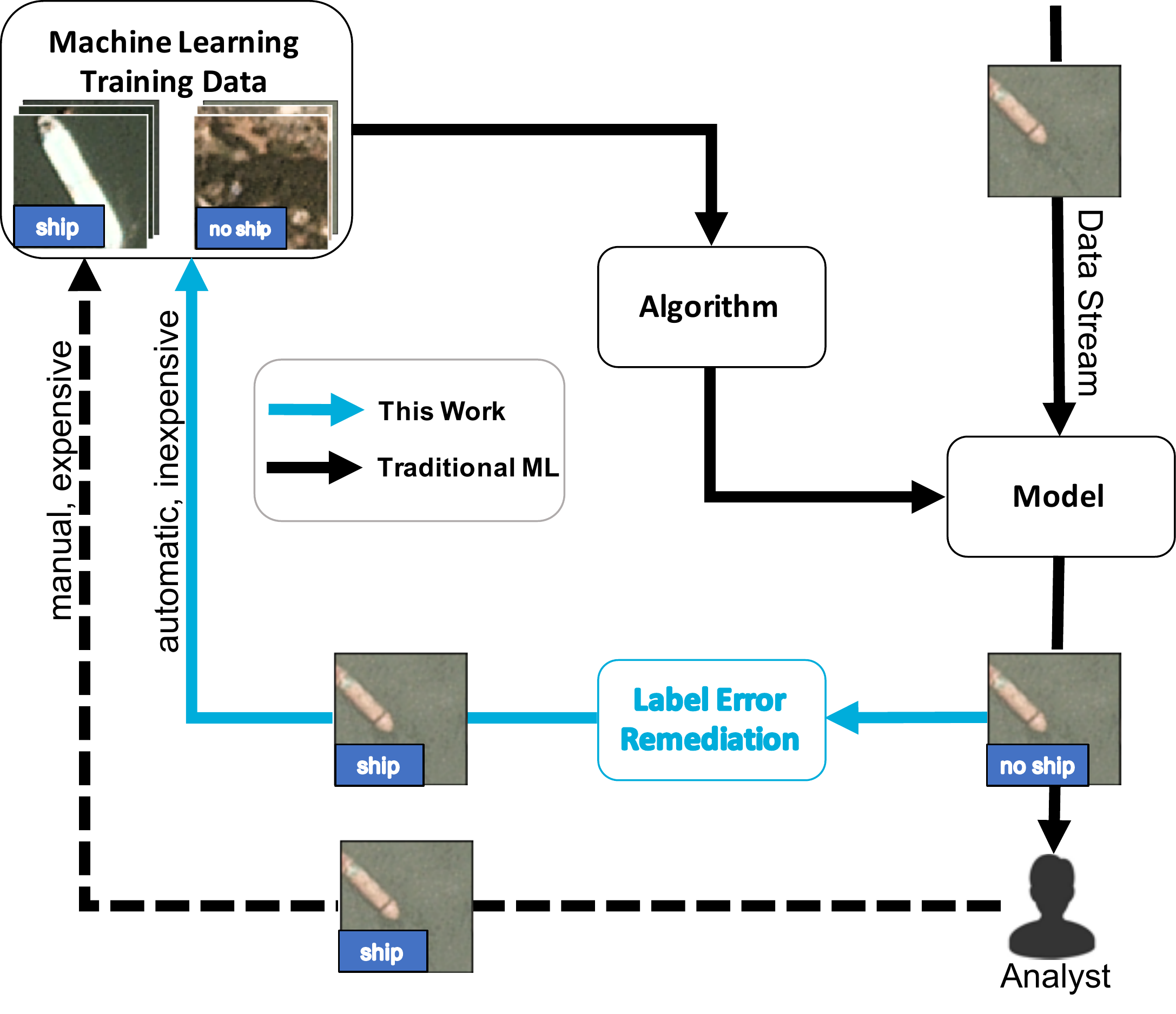}
  \caption{SUMER versus traditional machine learning for an example
  ship detection problem}
  \label{fig:metis}
\end{figure}

Self-Updating Models with Error Remediation (SUMER) is concerned with autonomously updating deployed machine
learning models as data naturally or adversarially drifts over time in
order to maintain desired performance. Figure \ref{fig:metis}
illustrates the continuous updating process employed by the SUMER
framework overlaid with a more traditional machine learning
deployment. The black lines represent a traditional machine learning process and the blue lines 
represent the SUM/SUMER framework. 

\renewcommand{\figurename}{Figure}

The main difference shown in Fig.~\ref{fig:metis} is the mechanism
used to keep a model up-to-date with the observed data. In traditional
machine learning, a human typically labels instances from the data
stream, which are then inserted into the training data and the machine
learning model is updated. This process is depicted with the dashed
line as it is often not done in practice due to the labor-intensive
cost associated with manual labeling. The SUMER framework proposes to
eliminate this bottleneck by using the current model's predictions as
labels after correcting any potential errors.

However, in order to fully implement the proposed SUMER framework,
access to the following resources are required:

\begin{itemize}
\item Training data - the set of annotated (i.e., labeled) data
  used to derive the machine learning model to perform the desired
  task. For example, in the ship detection example, the set of images
  and corresponding label for each image (i.e., ``ship'' or ``no
  ship'') constitute the training data. 
  This data will be used
  to develop the self-updating and label correction
  models.
\item Machine learning algorithm - the algorithm and
  implementation used to induce a model that can be
  deployed. 
\item Access to data stream - samples of data from the deployed
  environment that were not observed in the training data. The data
  stream should be in the same format as the training data (e.g., the
  same feature space) or convertible to the same format. For
  experimentation, ground-truth labels for these samples would also be
  desirable to evaluate performance.
\item Access to production system - a way to deploy the SUMER
  framework and ``hook'' into an existing production system will also
  be necessary.
\end{itemize}

A checklist of currently defined requirements will be discussed in
Sec.~\ref{sec:compare}.

\subsection{Relevant Machine Learning Scenarios}
SUMER is an amalgam of several machine learning scenarios in an
attempt to reduce or eliminate common issues that arise when utilizing
machine learning in practical environments. A list of relevant
scenarios and their definitions follows (along with some relevant
references in open research):
\begin{itemize}
\item \textbf{Active Learning} \cite{Settles09} - exploit what the model
  thinks that it doesn't know. This attempts to reduce the number of
  labels needed to update a model by allowing the model to query an
  oracle, usually a human expert, for labels on selected samples from
  the data stream.
\item \textbf{Concept Drift Detection} \cite{Gama14} - detect a shift in
  the distribution of the data, which is typically detrimental to
  model performance.  This could be a change in the data associated
  with known classes or the introduction of one or more previously
  unknown classes.
\item \textbf{Domain Adaptation} \cite{Zhang13}- adapt a model to a
  distribution shift between the training data used to induce a model
  and the data stream to which the model is applied. This typically
  assumes that the features used in the training data and data stream
  are equivalent (i.e., the input(s) to the model will be the same).
\item \textbf{Feature Augmentation} \cite{Crussell15, Daume10} - add
  or modify the features (i.e., input(s)) that the model uses to make
  its predictions in order to improve performance on the desired
  task. For example, new features can be created based on the outputs
  of auxiliary models (e.g., outlier detection).
\item \textbf{Learning with Label Noise} \cite{Natarajan13, Scott13} -
  induce a well-performing model given that some of the data is known
  to be mislabeled. This can be done by detecting potential errors and
  removing those samples, changing their labels, or by weighting those
  samples accordingly.
\item \textbf{Model Shift Detection} \cite{Raeder09} - determine when
  a deployed model is not performing as expected, but do not attempt
  to correct performance. Typically accomplished by monitoring the
  output of the model and detecting shifts in its distribution.
\item \textbf{Semi-supervised Learning} \cite{Zhu05}- exploit what the
  model thinks it knows. That is, assume that samples in the data
  stream are correctly labeled if the model is confident in its
  predictions. As in active learning, the goal is to reduce the number
  of labels needed from an oracle, which is an expensive process.
\end{itemize}

\subsection{Comparison of Learning Scenarios}
\label{sec:compare}
As the aforementioned learning scenarios are related, their
requirements can be similarly defined. Table \ref{table:compare} lists
the necessary requirements for each of the scenarios.  By determining
what is available for a potential application and comparing with
Tab.~\ref{table:compare}, the matching scenarios could potentially be
utilized. All that remains would be to decide if the problem solved by
the matching scenarios is of utility to the transition partner before
moving forward.

In summary, although as defined SUMER requires access to several
resources to be effective, the framework can be adapted to solve a
large variety of related problems, based on the needs of the
application and what can currently be provided.

\begin{table}[ht]
\caption{Comparison of various relevant machine learning scenarios to SUMER}
\label{table:compare}
\begin{center}
\resizebox{\textwidth}{!}{ 
\begin{tabular}{|c||c||c|c|c|c|c|c|c|c|}
  \hline
  & \rot{\textbf{SUMER}} & \rot{\textbf{Traditional ML}} & \rot{\textbf{Active Learning}} & \rot{\textbf{Concept Drift Detection\ }} & \rot{\textbf{Domain Adaptation\ }}
  & \rot{\textbf{Feature Augmentation}} & \rot{\textbf{Learning w/ Label Noise}} & \rot{\textbf{Model Shift Detection}} & \rot{\textbf{Semi-superivsed Learning}}  \\
        \hline
    Training Data   & \OK & \OK & \OK   & \OK & \OK & \OK & \OK & &   \OK   \\
    \hline
    Data Stream    & \OK & \OK & \OK & \OK & \OK & \OK & \OK &&  \OK \\
    \hline
     Output of Deployed Model on Data Stream  & \OK & \OK   &  \OK  &   & & & & \OK & \OK  \\
     \hline
     Mechanism to Update Deployed Model  & \OK & \OK &  \OK  & & \OK & & \OK & & \OK \\
     \hline
    Self-prediction Mechanism to Label Data Stream & \OK &  &    &  &  & & & & \OK \\
    \hline
    Label Error Detection / Correction & \OK &  & & & &  & \OK &  &    \\
    \hline
     Analyst / Oracle to Label Data Stream &  &  \OK  & \OK  & &   &    &    & &     \\
     \hline
    \end{tabular}
  }
 \end{center}
\end{table}

\subsection{Synopsis}


In order to advance an autonomous machine learning framework that
self-adapts to changing environments, the most immediate problem that
was identified was that of error propagation in SUMs. If models are to
be self-updated in practical, deployed environments with minimal human
intervention, then these models need to be able to identify and
self-correct mistakes. To successfully address this problem, the task
was divided into two main thrusts: 1) demonstrate the benefits of
self-updating and 2) demonstrate the benefits of self-updating with
error remediation\footnote{We use the terms \emph{error remediation}
and \emph{label correction} interchangeably.}

The remainder of the paper is as follows. Section~\ref{sec:problem}
defines the problem and provides more intuition surrounding
it. Section~\ref{sec:related} identifies and summarizes open research
that is relevant to this work. It also provides an initial taxonomy
for SUMs and remediation techniques. Section~\ref{sec:experiments}
describes the experiments that were performed during the course of
this research in order to understand SUMs and SUMER. This section
also describes some issues that were identified with utilizing these
models in practical environments. Section~\ref{sec:conclusion}
concludes the report and provides proposed next stages for this
research.

%% file: tex/problem.tex
\begin{figure}[!h]
    \centering \subfloat[]{{\includegraphics[width=.25\textwidth]{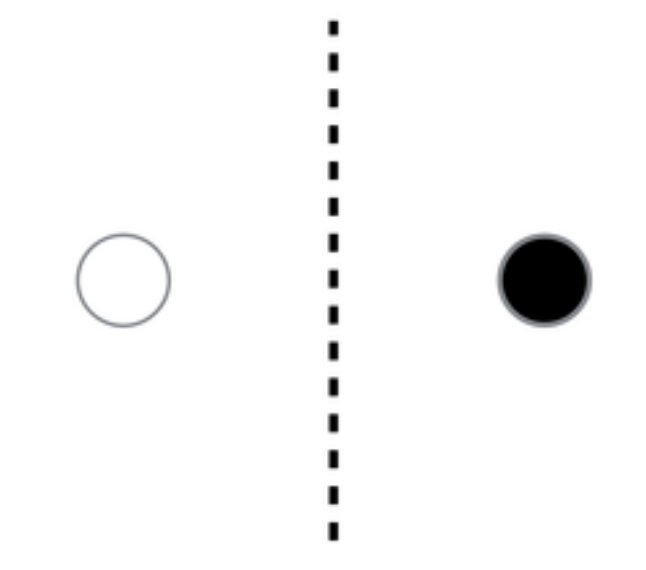} \label{fig:example-1a}}} \qquad \hspace{3cm} \subfloat[]{{\includegraphics[width=.31\textwidth]{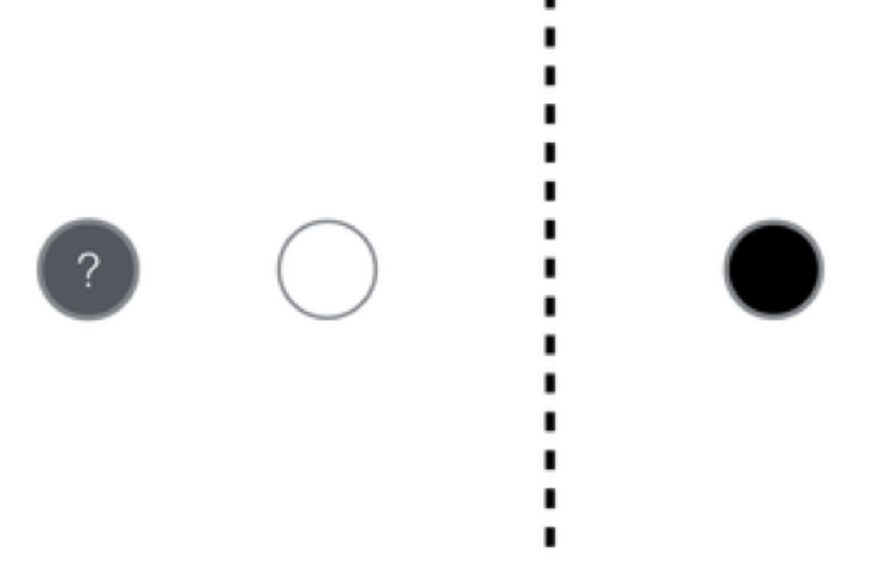} \label{fig:example-1b}}} \caption{Illustrative
    example of traditional machine learning (a) Given two labeled data
    points (with or without additional unlabeled data points), a
    decision boundary is inferred. (b) When a new unlabeled data point
    is presented, lacking any additional information, the inferred
    decision boundary is the same, dictating the prediction of the new
    point\cite{Ibanez19}.}  \label{fig:example1}
\end{figure}

The problem of self-updating models can be seen in the following
example.  Assume that the task is to differentiate between black and
white points.  In this setting, we are assuming that more data will
arrive as the model is deployed.  Given a labeled point from each
class, a decision boundary can be inferred as illustrated in
Fig.~\ref{fig:example-1a}.  Given a new point, the gray point in
Fig.~\ref{fig:example-1b}, how should that new point be labeled?
Without further context, the best that the model can do is categorize
the point as belonging to the ``white'' class.

\begin{figure}[!h]
    \centering
    \subfloat[]{{\includegraphics[width=.3\textwidth]{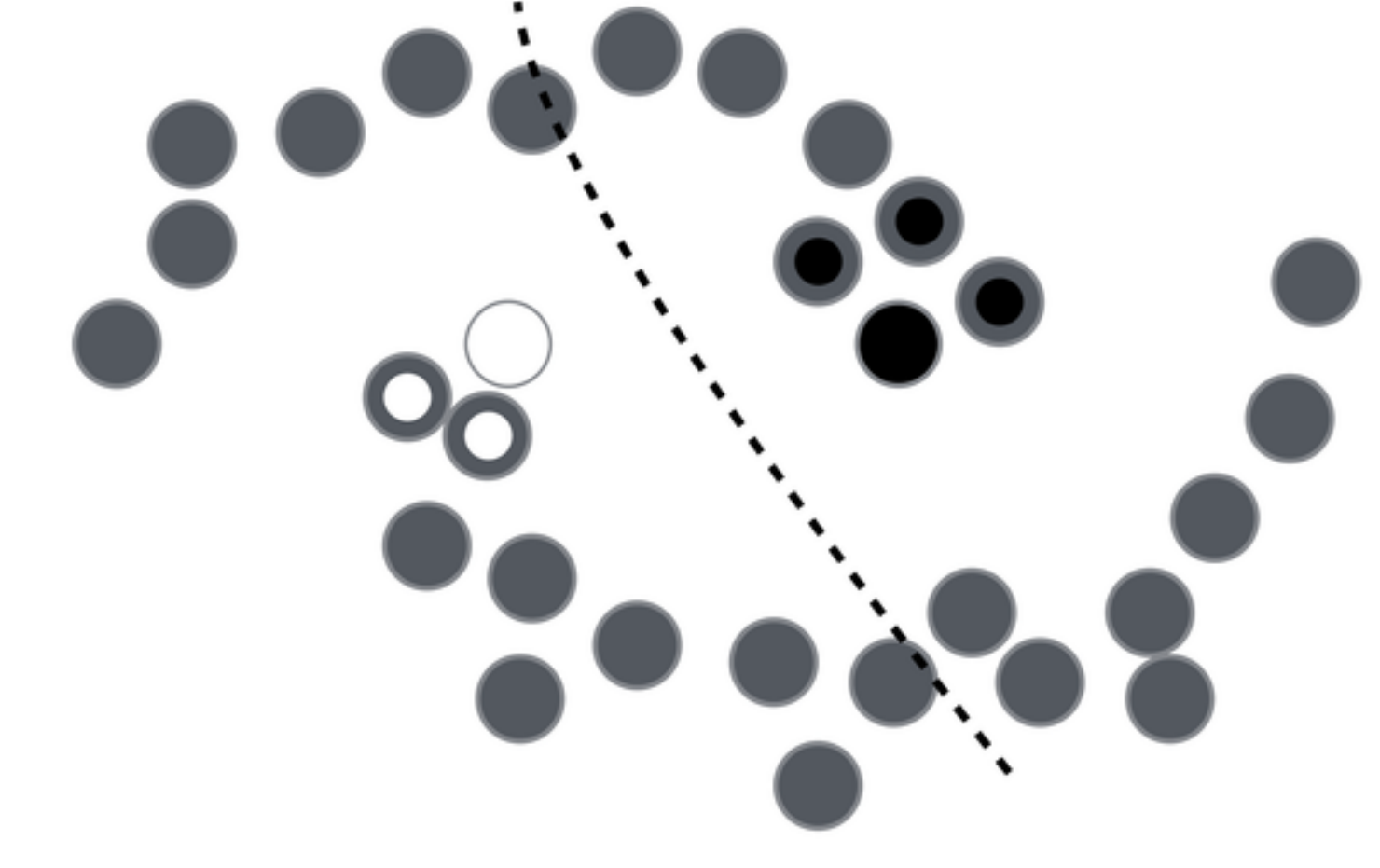} \label{fig:conceptual-2a}}}
    \qquad
    \subfloat[]{{\includegraphics[width=.3\textwidth]{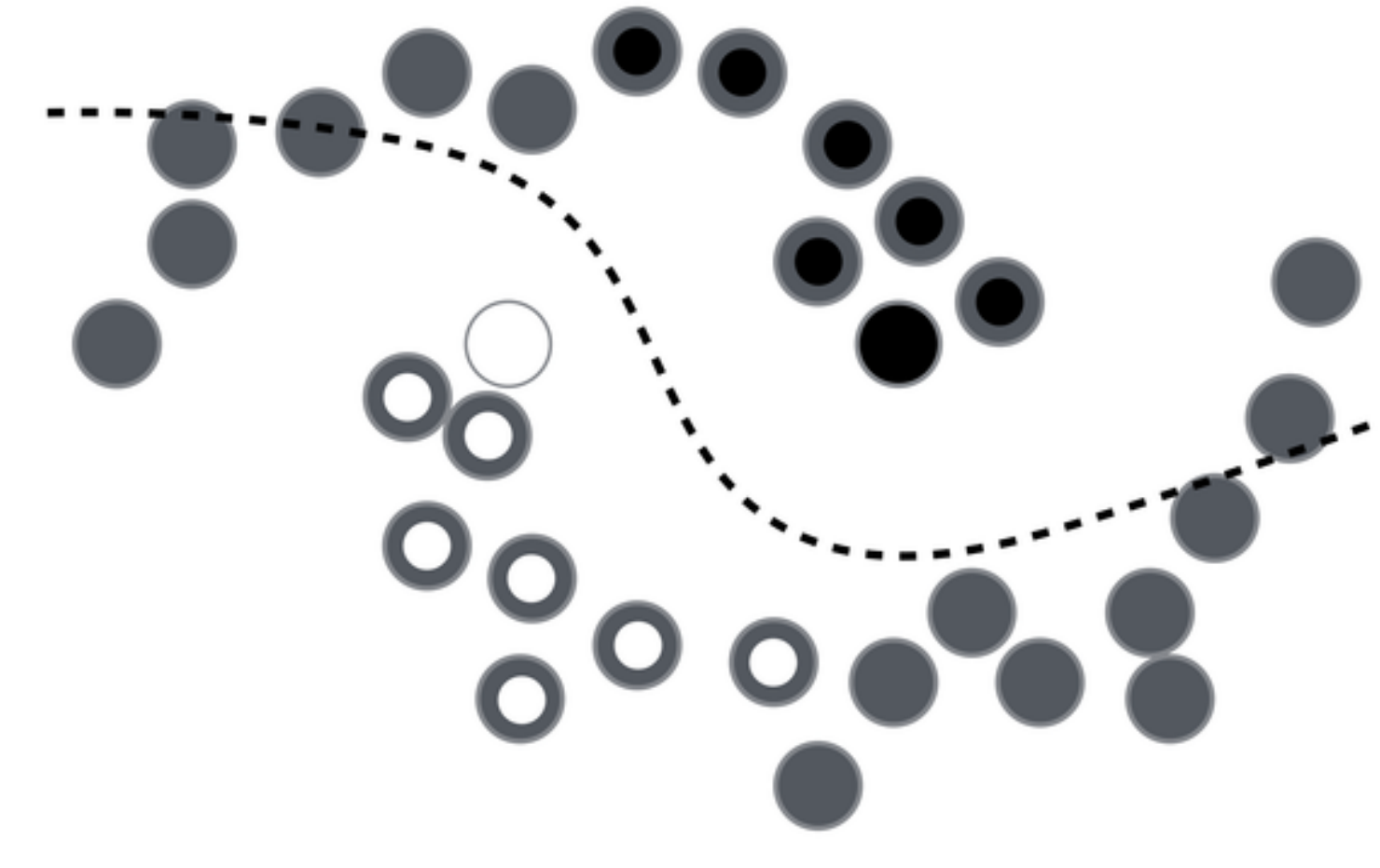} \label{fig:conceptual-2b}}}
    \qquad
    \subfloat[]{{\includegraphics[width=.3\textwidth]{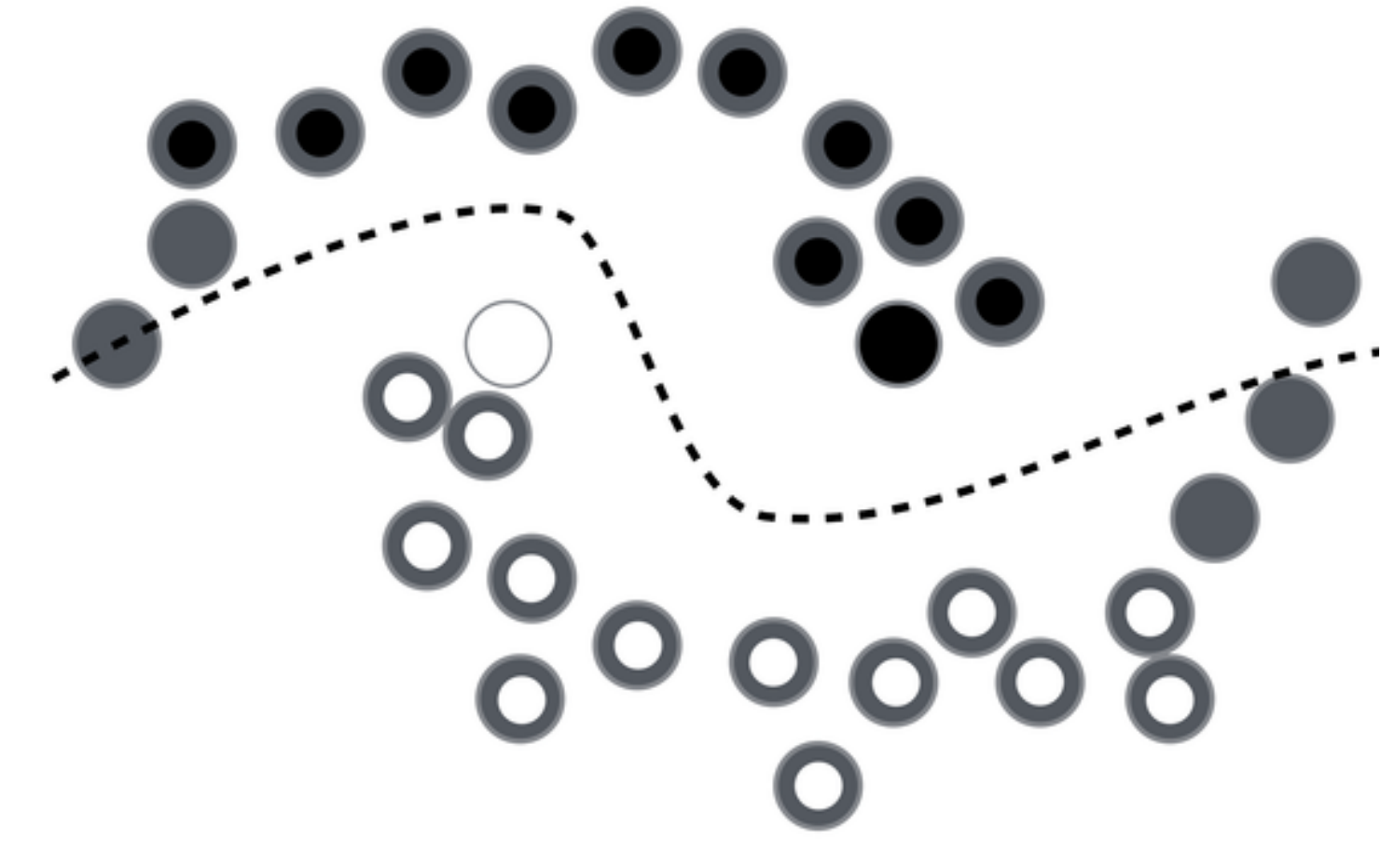} \label{fig:conceptual-2c}}}
    \qquad
    \subfloat[]{{\includegraphics[width=.3\textwidth]{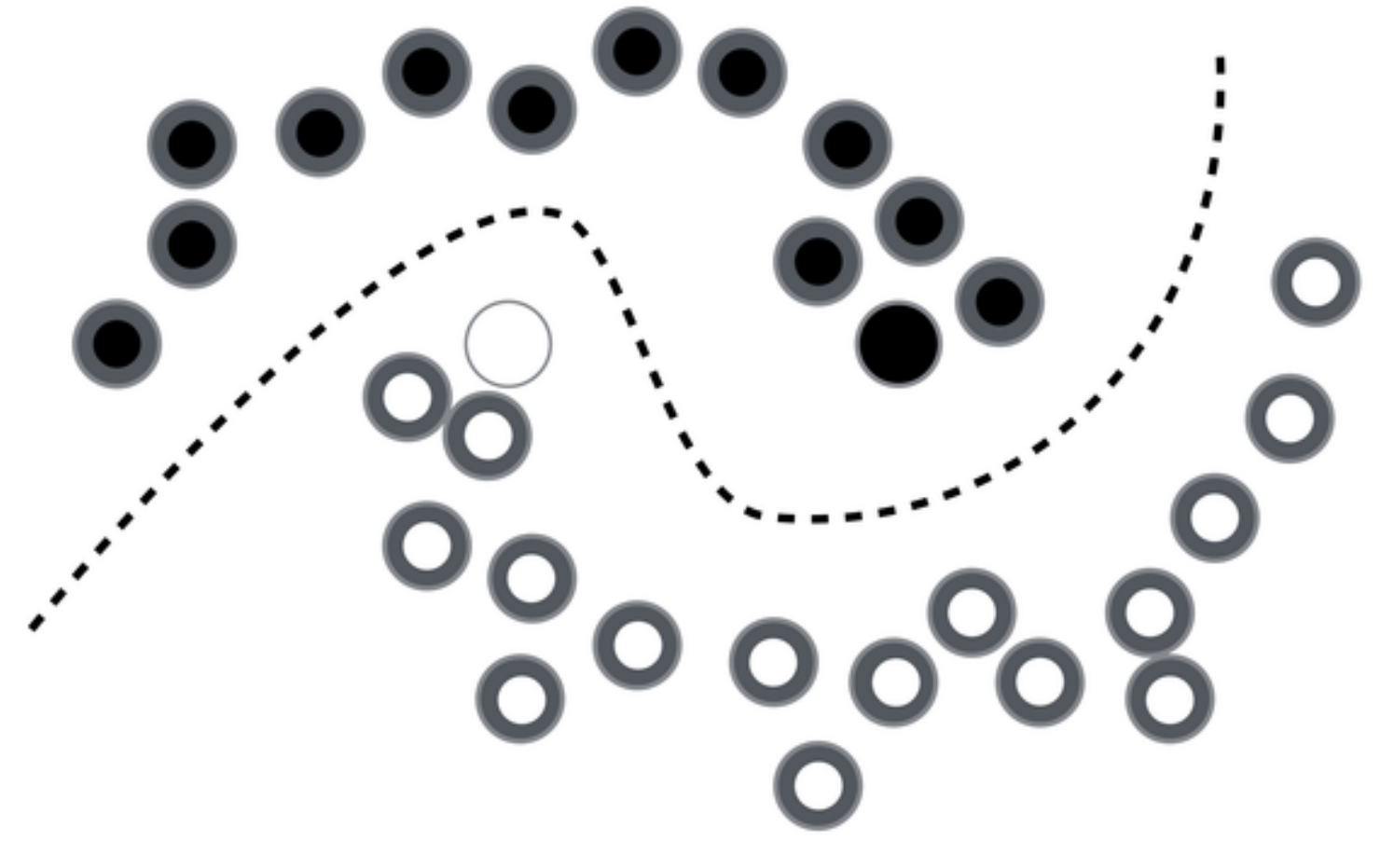} \label{fig:conceptual-2d}}}
    \caption{Illustrative example of iteratively labeling unlabeled data points and retraining an algorithm to infer a decision boundary\cite{Ibanez19}.}
    \label{fig:example2}
\end{figure}

Now, under the assumption that more unlabeled data is constantly being
observed, that data can be leveraged to update the model as
illustrated in Fig.~\ref{fig:example2}a-d.  This examples illustrates
what happens if the data is updated in batches based on intermediate
self-updating of the nearest points.  If all of the data were labeled
at once based on proximity to the original labeled data points, then
the classification boundary would have appeared similar to the
original decision boundary in Fig.~\ref{fig:example-1a}.

\begin{figure}[!ht]
\centering
\includegraphics[width=.4\textwidth]{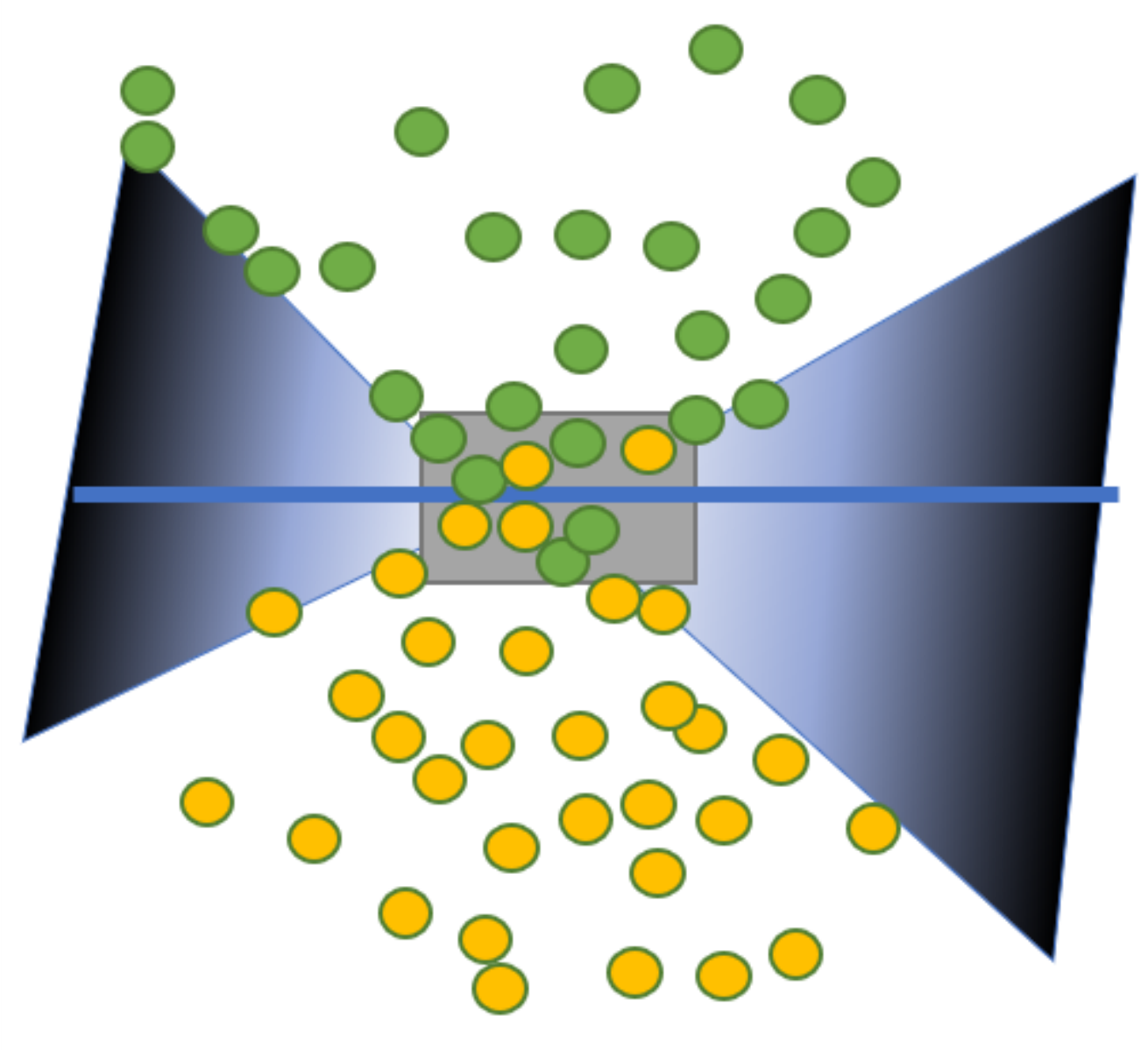}
\caption{Notional illustration of uncertainty of a machine learning
model discriminating between green and yellow. The gray box represents
uncertainty in the feature space as the classes overlap. The
blue-black gradient triangles represent uncertainty due to a lack of
training data in that portion of the input space.}
\label{fig:concept}
\end{figure}

By labeling the data in a strategic manner, the correct decision
boundary could be inferred.  Self-labeling will often use the labels
from a model about which it is most confident.  Confidence in many
models is estimated by the distances from a decision boundary.
However, in examining the model, there are different areas of
uncertainty that should be taken into account when performing
self-labeling as illustrated in Fig.~\ref{fig:concept} representing a
decision boundary (the solid blue line) between the green and yellow
class.  These are represented by the gray box in the middle of the
decision boundary and the blue-black gradient extending toward each
side.  The source of uncertainty of the gray box is from the
overlapping points from differing classes.  The source of uncertainty
in the blue-black gradient triangles is because there is a lack of
data on those areas.  Thus, even if the model is confident about a
prediction (e.g. the data point is far away from the decision
boundary), if the data point is not represented by the training data,
the confidence should be low.

The end goal, therefore, is to autonomously update a learned model
using its own output on new data points where: 1) the training data
does not cover and 2) there is high confidence that the model can
extrapolate or generalize to that new area.  In the streaming sense,
we aim to remediate incorrect predictions so that errors are not
propagated in future iterations and we would also like to detect
concept drift and novel concepts.  The novelty in the proposed
research lies in the fusion of multiple algorithms in deployed
environments that adapt over time. Most of the open research assumes a
fixed dataset or does not address the issue of multiple rounds of
self-updating. However, there are several outstanding issues not
considered during the course of this research. For example, this work
does not attempt to address drift in the underlying concepts
associated with the learning task. Section~\ref{sec:conclusion}
provides more detail on remaining research gaps and potential
solutions.

%% file: tex/experiments.tex
In this section, we will describe the various datasets that were used
for experimentation, along with the experimental set up and results
obtained for this data. Additionally, we will discuss some of the
conclusions that we have drawn from the outcomes of our
experiments. Finally, we will describe some potential issues that were
discovered over the course of our research involving the use of
self-updating models and label error remediation in practice.

\subsection{Datasets}

Before discussing the various experiments that were performed, it is
useful to understand the datasets that were used. A few datasets were
utilized in order to ensure that any conclusions obtained from our
experiments would generalize to similar machine learning problems.

\subsubsection{Synthetic framework}

In machine learning research, it is often useful to perform
experiments on synthetically-generated data. By using synthetic data
with known properties, it allows for the understanding of algorithms
without having to account for the various complexities associated with
real-world data, which are often unknown or difficult to
summarize. Therefore, a synthetic data generation and label noise
insertion framework was developed for experimentation purposes. This
framework proved very useful for determining potential practical
issues associated with self-updating models and label error
remediation techniques.

\begin{figure}[!ht]
    \centering
    \subfloat[Two Gaussians]{{\includegraphics[width=.4\textwidth]{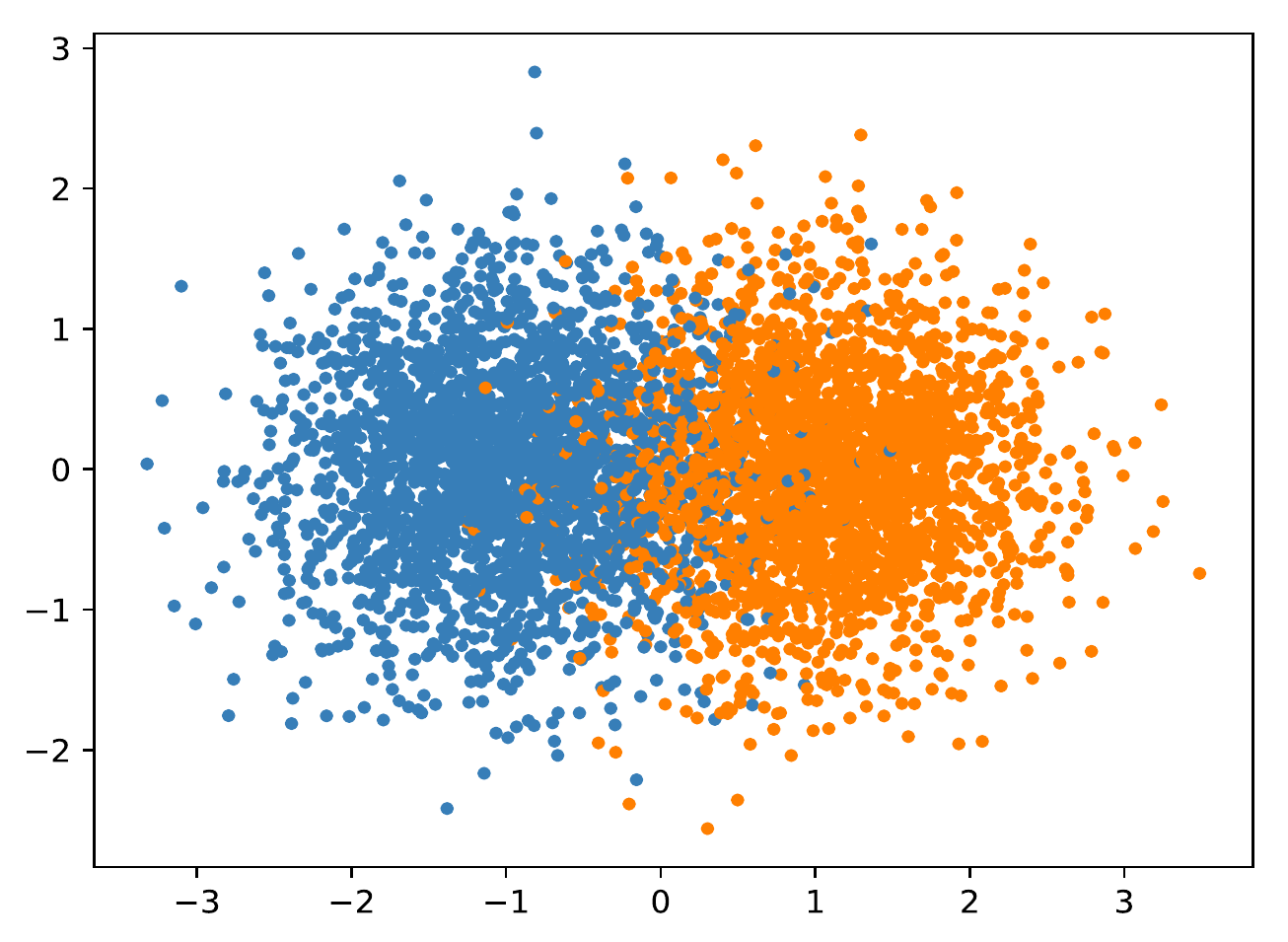} \label{fig:two-gauss}}}
    \qquad
    \subfloat[Two Moons]{{\includegraphics[width=.4\textwidth]{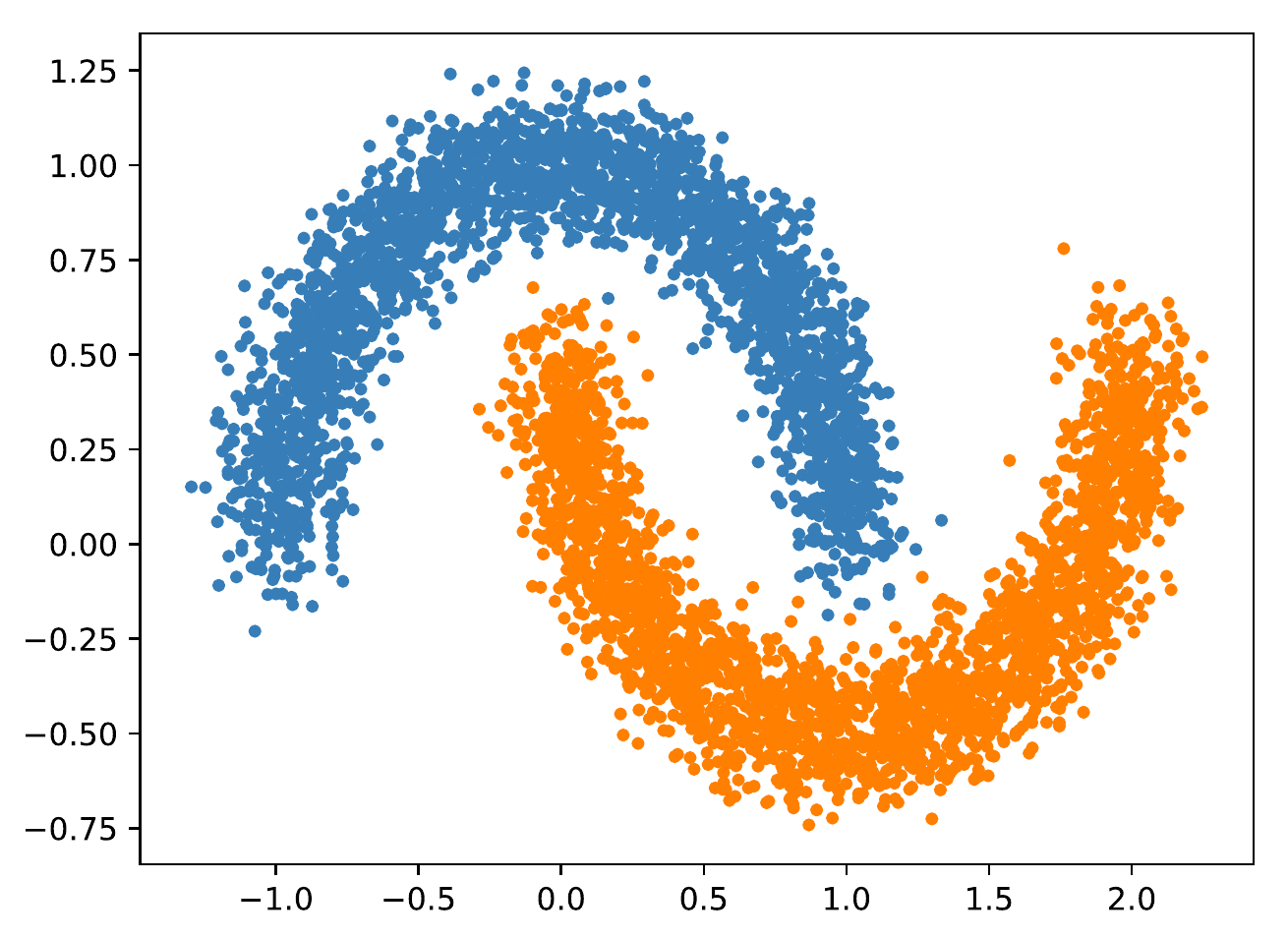} \label{fig:two-moons}}}
    \caption{Examples of synthetically-generated data for two-class (binary) machine learning problems}
    \label{fig:synthetic-data}
\end{figure}

Figure~\ref{fig:synthetic-data} shows data that was generated by the
developed framework for two different binary machine learning
problems. Here, the goal of the model is to categorize a data point as
blue or orange. The features for each point are simply the coordinates
in two-dimensional space. The first example, shown in
Fig.~\ref{fig:two-gauss}, generates data from two separate
multivariate Normal distributions.  The second example, shown in
Fig.~\ref{fig:two-moons}, generates data points for each of two
overlapping ``moons''. In this case, the optimal classifier is
nonlinear.

\subsubsection{Ships in satellite imagery}
\label{data:kaggle}

\begin{figure}[!ht]
    \centering
    \subfloat[No Ship]{{\includegraphics[width=.5\textwidth]{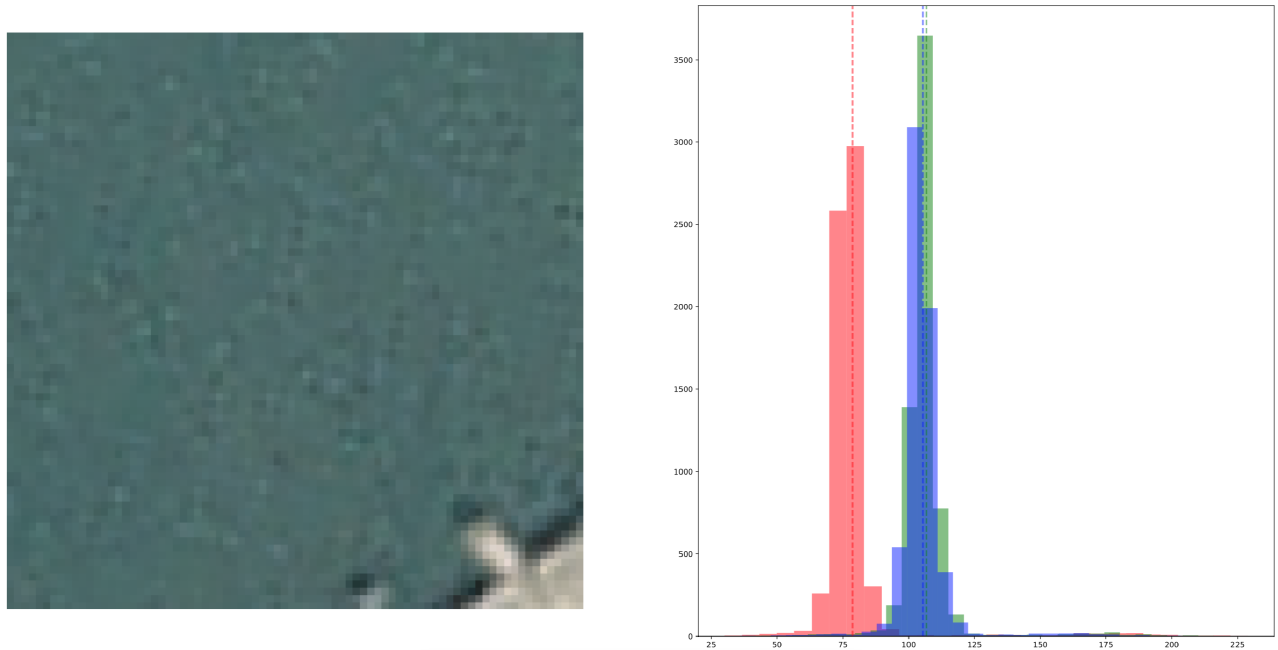} }}
    \qquad
    \subfloat[Ship]{{\includegraphics[width=.5\textwidth]{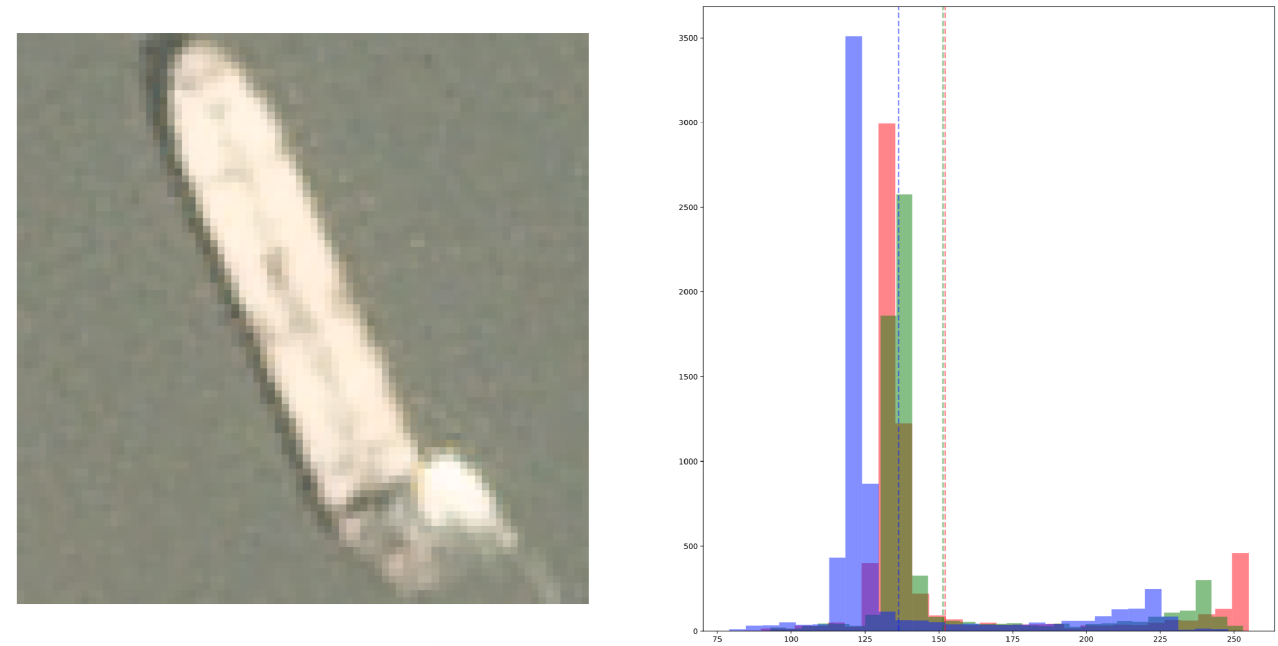} }}
    \caption{Examples of images for the ship detection problem and
      their respective RGB histograms}
    \label{fig:ships-data}
\end{figure}

The Kaggle ``Ships in Satellite Imagery'' dataset
\cite{Kaggle19} is comprised of a collection of $80\times80$
red-green-blue (RGB) chips that were extracted from satellite images
provided by Planet\footnote{\url{http://www.planet.com}}. The goal of
this dataset is to induce a model that can detect whether or not a
given chip contains a ship (i.e., predict ``ship'' or ``no
ship''). The features for this problem are the pixel intensities in
the red, green, and blue channels for each of the pixels, which are
represented as integers in $[0, 255]$. Given that there are three
channels and $80\times80$ pixels, the resulting feature space is quite
large and each chip is represented by 19,200
integers. Figure~\ref{fig:ships-data} shows an example from each of
the two classes, along with the histograms of pixel intensities.

Originally, there were 1,000 examples of ships and 3,000 examples of
non-ships, for a total of 4,000 image chips in the dataset. However,
the non-ship class is composed of 1,000 land-cover chips, 1000 partial
ships, and 1,000 commonly misclassified chips that do not contain
ships. We consider the 1,000 partial ships and 1,000 commonly
misclassified chips to be associated with concept drift and were
therefore left out of some of the experiments. For relevant
experimental results, we will explicitly mention when these chips were
held out.

Additionally, for future experiments involving concept drift, it would
be useful to re-label the partial ships to be a part of the ``ship''
class. In this instance, this re-labeling procedure would be of more
interest from a practical perspective as the chip still contains a
part of a ship. However, as concept drift was not considered for our
initial experiments, this distinction is not required.

\subsection{Experiments on Synthetic Data}

\begin{figure}[!ht]
    \centering
    \subfloat[Two Moons with only a few samples labeled]{{\includegraphics[width=.4\textwidth]{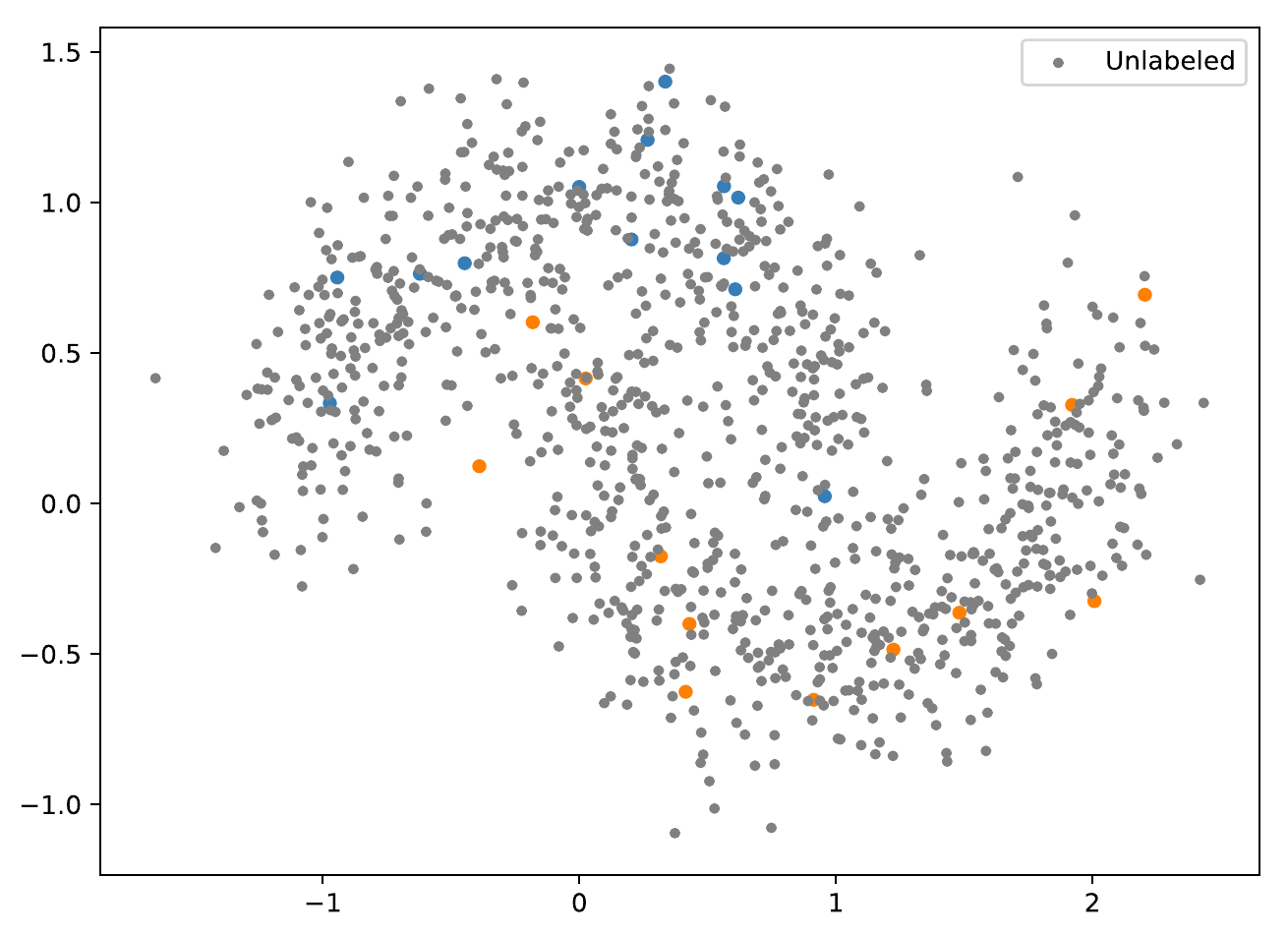} }}
    \qquad
    \subfloat[Resulting SUM predictions]{{\includegraphics[width=.4\textwidth]{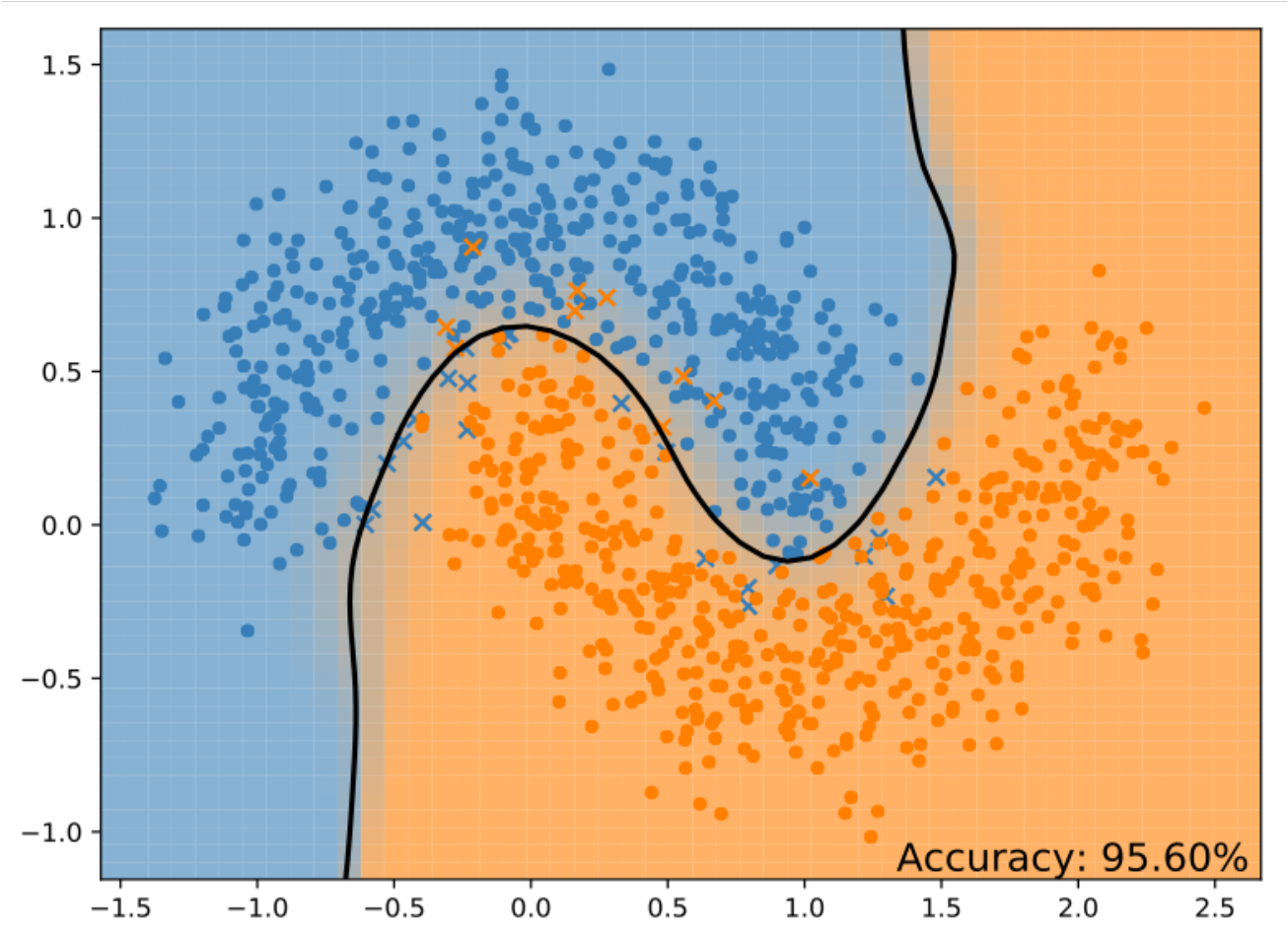} }}
    \caption{Illustration of SUMs on Two Moons dataset}
    \label{fig:synthetic-sums}
\end{figure}

The first experiment aimed to demonstrate the benefit of SUMs on
synthetic data, specifically the Two Moons
dataset. Figure~\ref{fig:synthetic-sums} shows the result of a SUM
applied to this data using label spreading \cite{Zhou03}. The left
side of the figure shows the data. The unlabeled data is shown by the
gray points and the labeled points are shown by the blue/orange
points. The right side of the figure shows the learned model. With an
accuracy of over $95\%$, it is clear that label spreading is able to
learn a good model with only a few labeled points. However, this
experiment only shows a SUM when all of the unlabeled data is known in
advance (i.e., a single round of self-updating).

\begin{figure}[!ht]
    \centering
    \includegraphics[width=.6\textwidth]{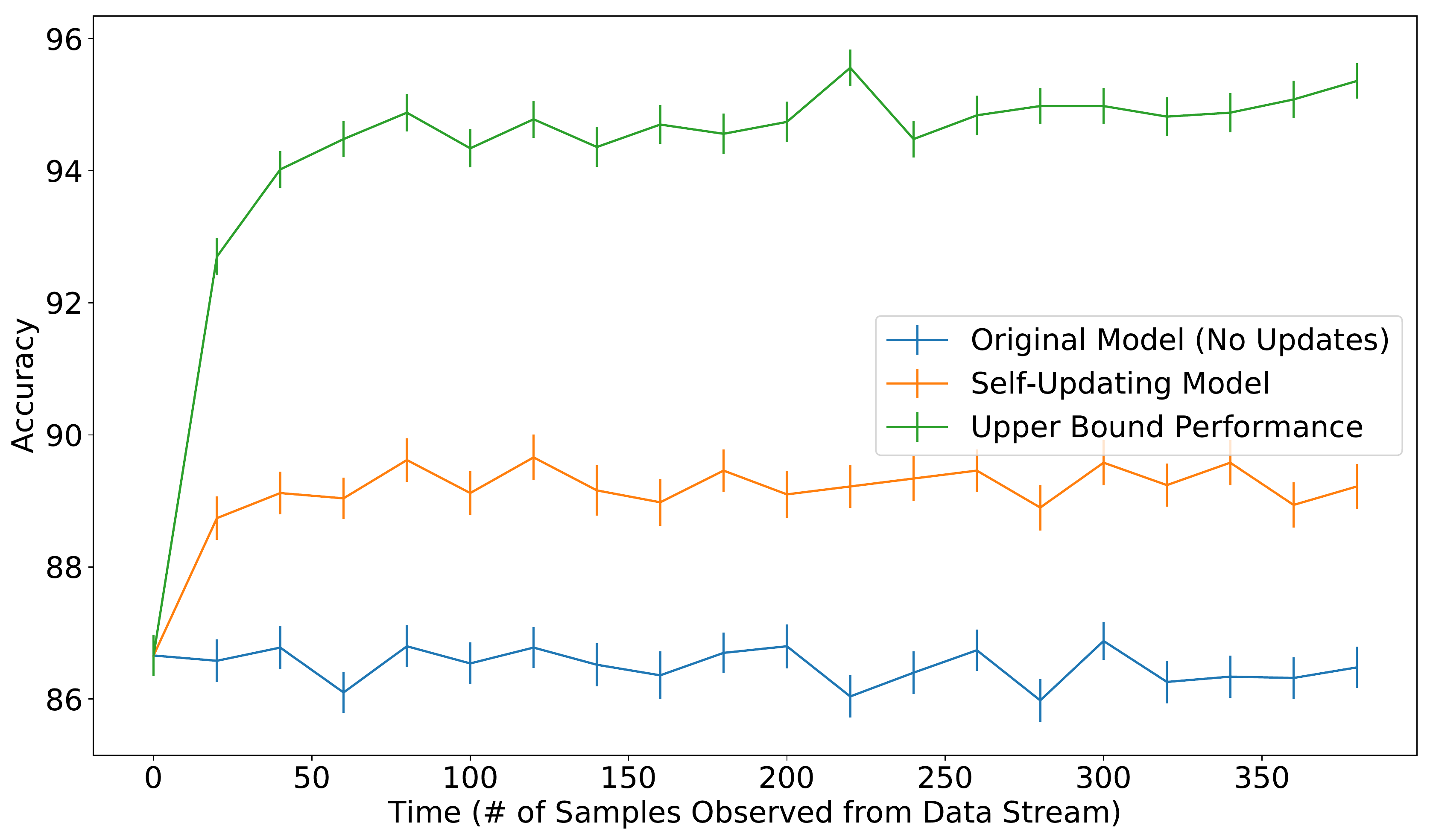} 
    \caption{Benefit of SUMs on a simulated stream over time}
    \label{fig:synthetic-sum-stream}
\end{figure}

Therefore, the next experiment investigated the performance of SUMs
over multiple rounds of
self-updating. Figure~\ref{fig:synthetic-sum-stream} shows the
performance of a SUM over a simulated stream. The algorithm is
provided with an initial labeled dataset as in the experiment with the
single round of self-updating. Initially, twenty points are labeled.
With only a single round of self-updating, all the remaining unlabeled
data is self-labeled by the model and then the model is
updated. However, for multiple rounds of self-updating, we simulate a
stream by continuing to draw new instances from the same distribution
and periodically updating the model with this new data. In this case,
the SUM provides a clear benefit of over $2\%$ in accuracy (the orange
curve) compared to an initial model that does not self-update (the
blue curve). The green curve provides an estimate of the upper bound
performance by allowing the algorithm to update with the correctly
labeled data (i.e., it has access to all of the labeled data at once).

\begin{figure}[!ht]
    \centering
    \subfloat[SUM]{{\includegraphics[width=.4\textwidth]{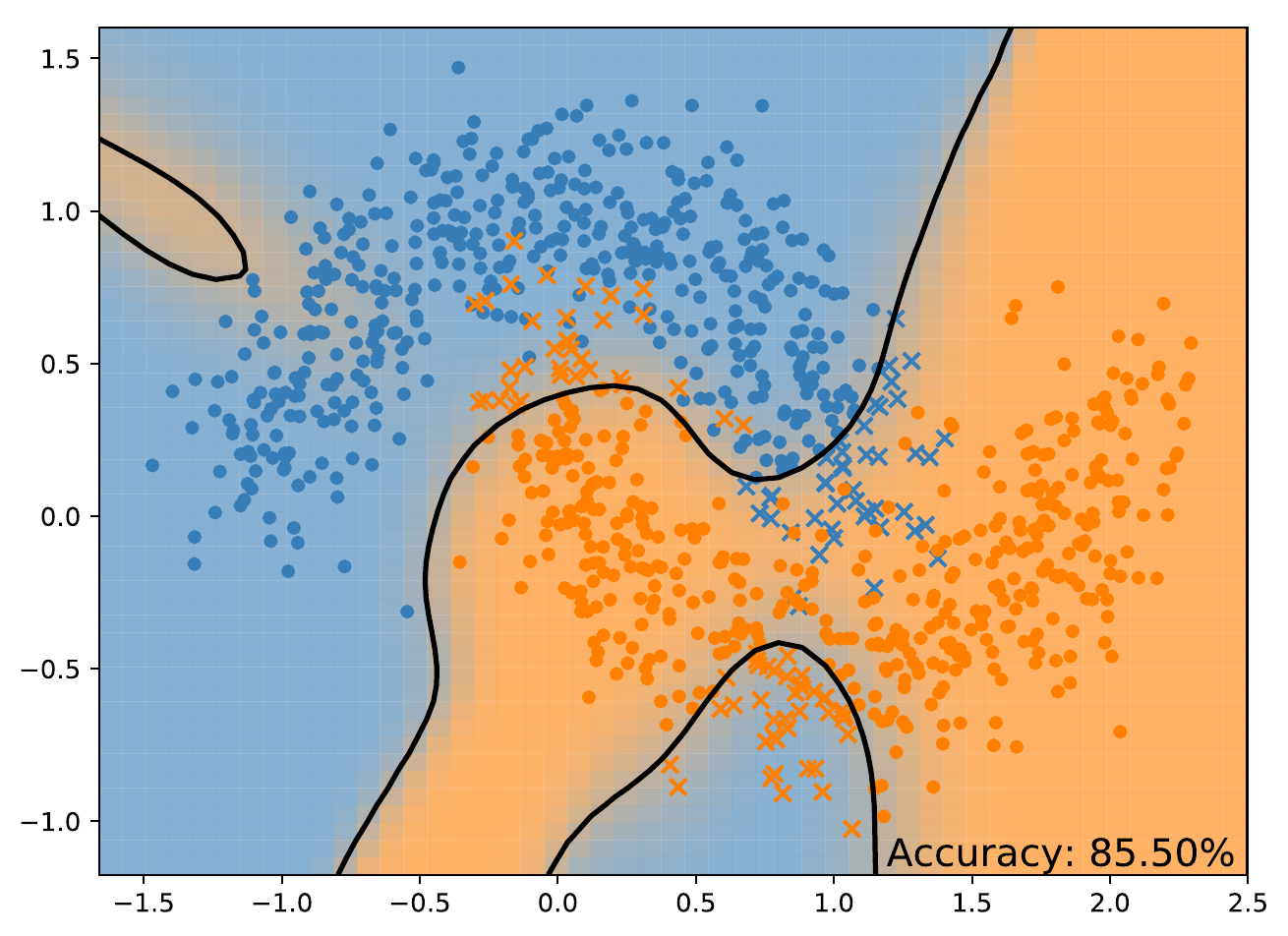} }}
    \qquad
    \subfloat[SUMER]{{\includegraphics[width=.4\textwidth]{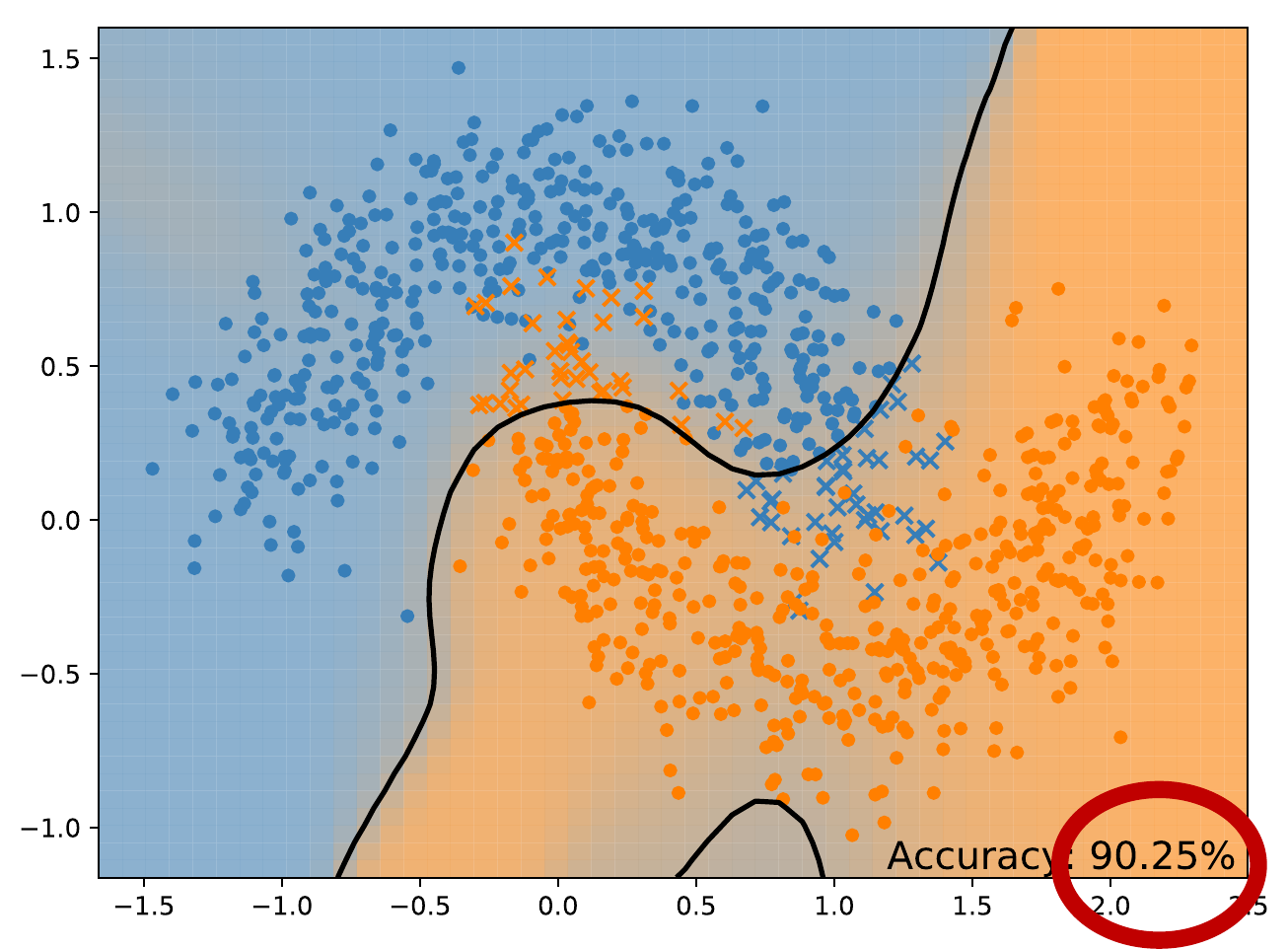} }}
    \caption{The benefit of SUMER in the presence of label noise}
    \label{fig:synthetic-sumer}
\end{figure}

The final experiment that was performed on the synthetic data involved
self-updating with error remediation
(SUMER). Figure~\ref{fig:synthetic-sumer} shows the models that were
learned from data that had $20\%$ label noise. The label noise was
introduced by randomly flipping labels. It can be seen that the SUMER
model provides an approximately $5\%$ increase in accuracy. Both
models utilized the label spreading algorithm, but in the case of
SUMER, a parameter was changed to allow provided labels to be
corrected, based on the optimization procedure derived by the
algorithm.

The aforementioned experiments demonstrate the benefit of SUMs/SUMER
on synthetic data, but these methods need to be vetted in more
realistic settings. Therefore, we performed similar experiments on
more realistic data.

\subsection{Experiments on Real-World Data}

\begin{figure}[!ht]
    \centering
    \includegraphics[width=.75\textwidth]{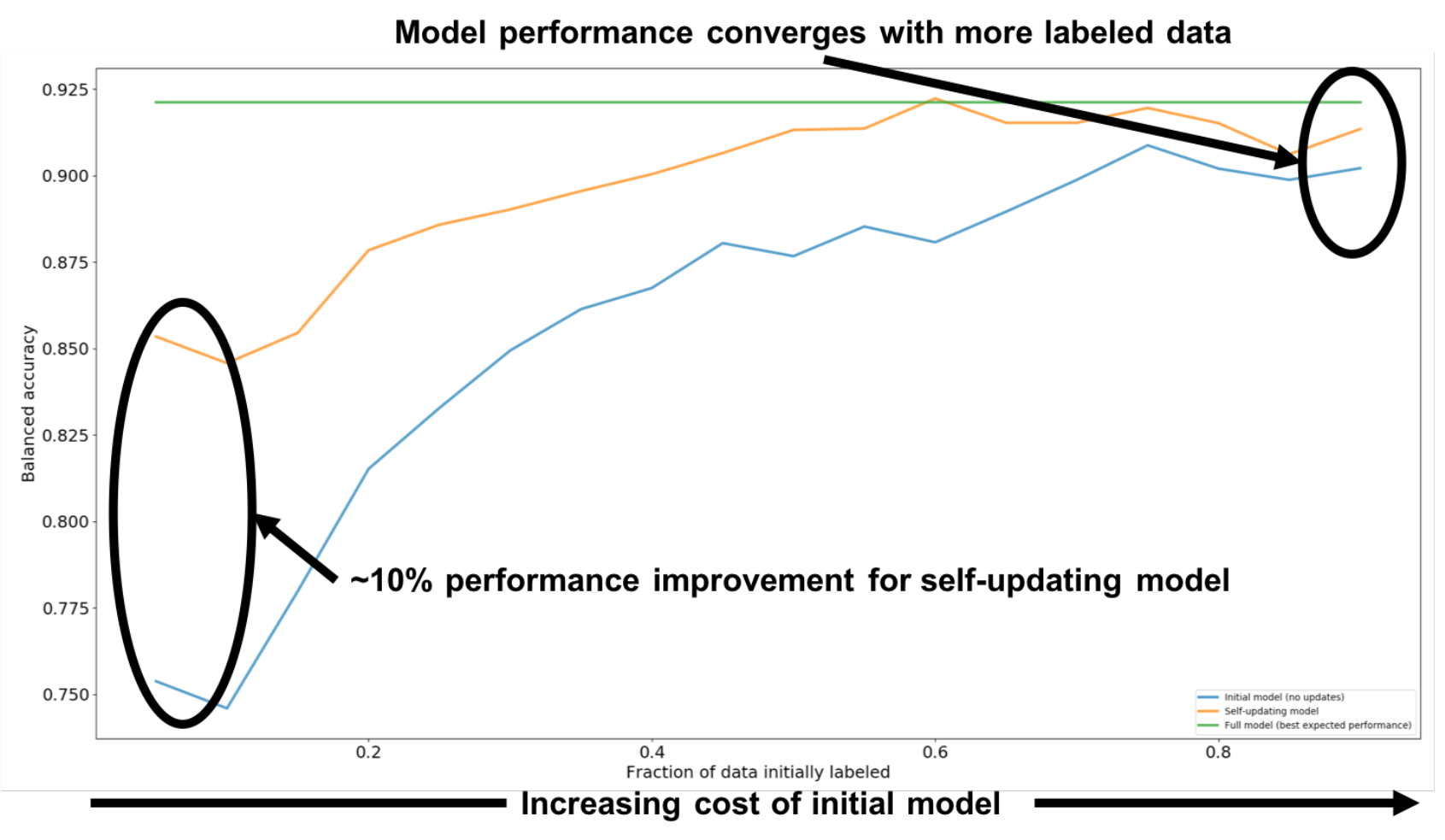} 
    \caption{Benefit of SUMs on Kaggle ships data}
    \label{fig:realistic-sums}
\end{figure}

As mentioned in Sec.~\ref{data:kaggle}, a realistic dataset
involving detecting ships versus landcover has been collected and
annotated. An initial experiment was conducted to demonstrate the
benefit of SUMs. For this experiment, the original, unmodified Kaggle
data was used (1000 ships and 3000 non-ships). For our model, we used
Random Forests \cite{Breiman01} in a self-updating
manner. Figure~\ref{fig:realistic-sums} shows the performance as a
function of the amount of data that is initially labeled. The $x$-axis
shows the fraction of data that has been labeled. The algorithm labels
the unlabeled data using the SUM and the performance accuracy is
estimated based on a held-out set of instances. When there is only a
small amount of data initially labeled ($\approx 5$-$10\%$), SUMs
provide an approximately $10\%$ increase in performance. As the amount
of initially labeled data increases, the performance converges to the
performance of the model with access to all of the labels, as
expected. However, in practice, the more data that is labeled, the
more labor-intensive the process is, which means that the model is
more costly to create.

The final experiment that was conducted involved evaluating the
performance of SUMER on a simulated stream of realistic data. In this
case, the data without drift was used (1000 ships and 1000
non-ships). However, $20\%$ label noise was injected into the initial
training data in order to evaluate the effect on the performance (the
true labels were used when estimating performance). The stream was
simulated by sampling (without replacement) instances from the full
set of data in predetermined windows. For this experiment, Random
Forests with Rank Pruning \cite{Northcutt17} was used for the SUMER
algorithm.

\begin{figure}[!ht]
    \centering
    \includegraphics[width=.75\textwidth]{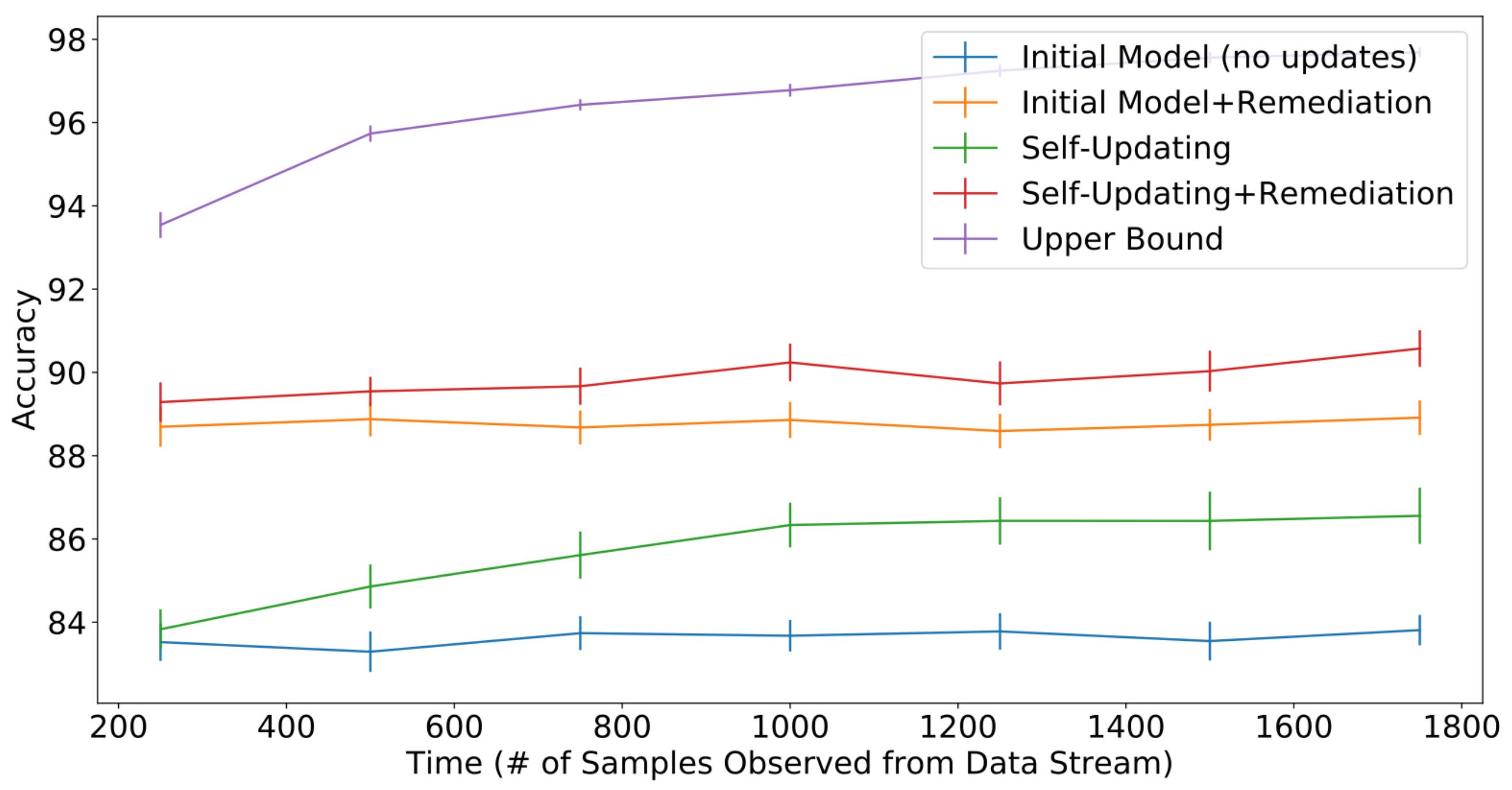} 
    \caption{Benefit of SUMER on Kaggle ships data with $20\%$ label noise}
    \label{fig:realistic-sumer}
\end{figure}

Figure~\ref{fig:realistic-sumer} shows the results of this
experiment. As expected, both SUM and SUMER perform better than an
initial model that does not update. SUMER comes the closest to the
upper bound performance (i.e., the model with full access to the
correct labels). It should be noted that the reason there is only a
small increase in accuracy observed between the initial model with
error remediation (the orange curve) and SUMER (the red curve) is
likely due to the model coupling problem, which will be discussed in
Sec.~\ref{subsec:remediation-issues}.

It seems that SUMs and SUMER are potentially very useful for
performing well in dynamic, label-constrained environments. However,
there are still some potential issues that should be discussed before
utilizing them in practice.

\subsection{Potential Issues with SUMs in Practice}
\label{subsec:self-updating-issues}

\begin{figure}[!ht]
    \centering
    \subfloat[Two Moons with only a few samples labeled]{{\includegraphics[width=.4\textwidth]{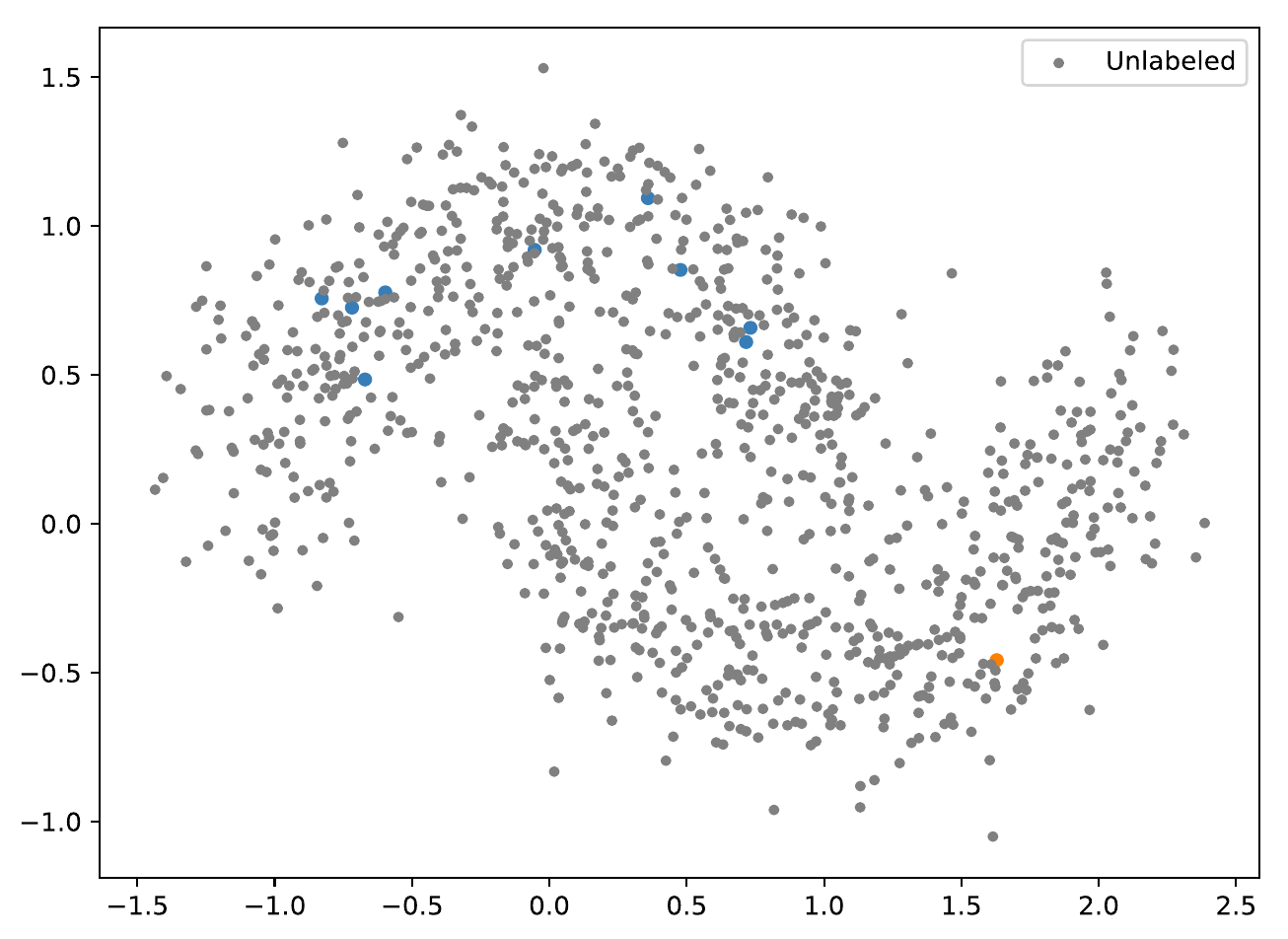} }}
    \qquad
    \subfloat[Performance degraded $\approx15\%$.]{{\includegraphics[width=.4\textwidth]{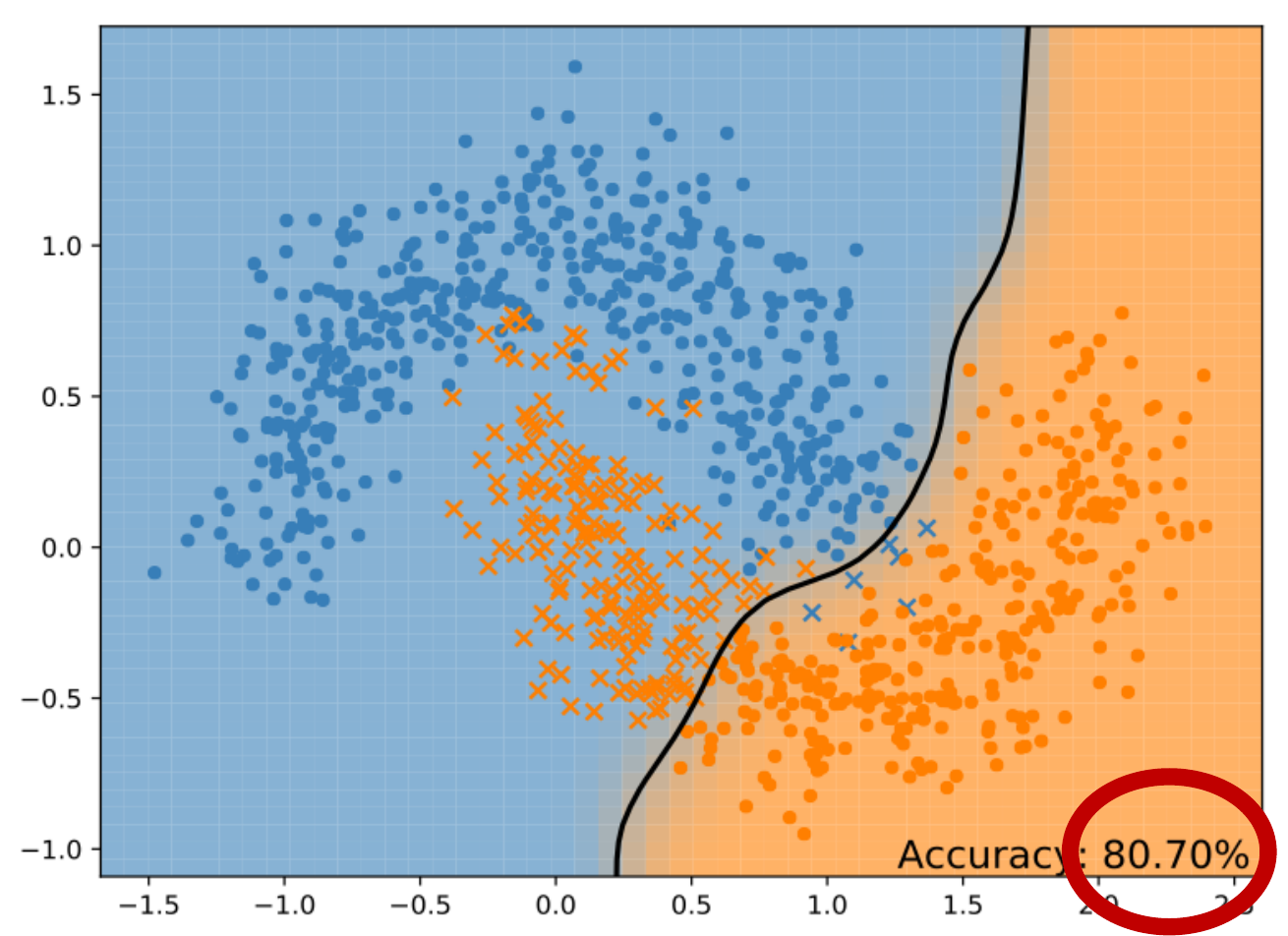} }}
    \caption{Illustration of initial label issue with SUMs}
    \label{fig:issue-initial-labels}
\end{figure}

One issue with using SUMs in practice surrounds the instances that are
initially labeled, which can drastically affect the overall
performance of the resulting
model. Figure~\ref{fig:issue-initial-labels} demonstrates this
issue. The data is sampled from the same distribution as in
Fig.~\ref{fig:synthetic-sums}, but the instances that are labeled are
different, which results in an approximately $15\%$ decrease in
performance. This problem is related to the problem of concept drift
\cite{Gama14}. Areas of the data distribution are not labeled, which
makes it difficult for the algorithm to determine a correct labeling
in those areas of the feature space. This difficulty, in turn, affects
the potential performance of the algorithm. The problem of concept
drift was not addressed during this research.

Additionally, the amount of data that is initially labeled will also
affect the performance increase observed by utilizing SUMs. When only
20 instances are labeled in the Two Moons dataset, the performance
benefit can be as much as $\approx 2\%$ as demonstrated by
Fig.~\ref{fig:synthetic-sum-stream}. However, as more labeled data is
initially provided to the algorithm, this performance gap closes.

\subsection{Potential Issues with Label Error Remediation in Practice}
\label{subsec:remediation-issues}

A potential issue may arise within the SUMER framework, which we refer
to as the model coupling problem.  To assist in the understanding of
the model coupling problem, consider that some label correction
techniques (see Sec.~\ref{sec:related-noise-rates}) estimate
class-conditional noise rates to determine remediation:
$P(\hat{Y}|Y)$.  This is the probability of guessing the wrong class,
$\hat{Y}$̂, when the actual class is $Y$.  If this is near-zero, then
no noise was detected and thus no remediation will occur.  This
manifests when labels inferred by label correction are
indistinguishable from SUM predictions (i.e., the models are coupled).
Notice that there is no dependence on the feature vector $X$.
  
For reference, a prediction model calculates $P(Y|X)$, which is the
probability of predicting class $Y$ given feature vector $X$.
Finally, consider a SUM whose behavior is described by
$P(\hat{Y}|Y,X)$.  This is similar to what a label correction
technique estimates except there is a dependence on the features.
When a SUM generates an incorrect prediction, which is also used as a
label for retraining, it is clear that the incorrect label is
dependent on the features because a feature vector was used to
generate the prediction.  This scenario corresponds to a small
emerging field in machine learning known as feature-dependent label
noise \cite{Scott19}.  Thus, model coupling can be described as a
disconnect between the label correction strategy, which assumes
feature independence, and a SUM, which is dependent on the features.
Possible solutions are to create an independent, auxiliary label
correction model; optimize loss that is immune to noise; and use
different views of the data if possible.  The last potential solution
is similar to the requirement for co-training to have different views
of the data.

%% file: tex/related.tex
In this section, we briefly describe some of the open academic
research that aligns with the framework defined by our research.
Additionally, this section is meant to provide references to other
potential algorithms and techniques that may be worth exploring in the
future.  However, please note that this list of related work is not
meant to be exhaustive.

At a high-level, the work presented here for SUMER fills a particular
niche that is only partially addressed in the open literature.
Specifically, many of the individual components are addressed in
isolation, but none of the current works put it all together in an
iterative fashion and take into account issues that arise from
multiple iterations of applying the various techniques or biases that
arise when using self-predicted labels.  Most of the work presented
here focuses solely on self-updating models or error and noise
remediation, for which we present related works in Secs.~\ref{sec:ssl}
and \ref{sec:noise}, respectively.  Other key areas include: 1)
uncertainty/trust in the model outputs such as active
learning \cite{Settles09} that encompasses strategies for querying a
human in the loop to obtain additional labels, output
calibration \cite{Hyams2017_Arxiv} that seeks to improve model
confidence, and uncertainty quantification in machine
learning \cite{Stracuzzi2018_SAND}; 2) understanding the
data \cite{Smith2014_MLJ}; 3) concept drift \cite{Gama14}; 4) and
online learning \cite{Saad1999}.  While substantial, this work only
scratches the surface of the possible avenues to explore.  Future work
will build upon this initial study and incorporate additional avenues
for research.

\subsection{Self-updating Models and Semi-supervised Learning}
\label{sec:ssl}
In general, self-updating models fall under the semi-supervised
learning paradigm in machine learning.  Semi-supervised learning
refers to techniques that have a set of data points that are labeled
and usually with a significantly larger number of data points without
labels.  For SUMER, we expand on this notion and assume that there
exists a set of labeled training data points to build a machine
learning model that will be deployed in a dynamic environment where it
will be exposed to large amounts of data that may not be represented
in the initial labeled training set.  Thus, self-updating will allow
the model to adapt to a richer set of data than what was available in
training.  In semi-supervised learning, many algorithms attempt to
assign labels to the unlabeled data points and then use the newly
labeled data to improve the training.

Most approaches differ in calculating the relationship between the
data points including: 1) clustering, 2) self-training, 3) multi-view
learning, and 4) self-ensembling.  Probabilistically, the methods
attempt to infer the probability $P(y|\mathbf{x})$ of a class $y$
given a data point ($\mathbf{x}$).  Early works clustered data
points \cite{McLachlan1975} \cite{Titterington1985}, examining how the
unlabeled data affects the shape and size of the clusters and using
Baye's theorem to approximate $P(y|\mathbf{x})$:
$P(y|\mathbf{x}) \propto P(\mathbf{x}|y)P(y)$.  Later work estimated
$P(y|\mathbf{x})$ directly using the predicted labels from models
trained on the labeled training set.  Each of the following
subsections will give an example of a few algorithms.
  
\subsubsection{Cluster-based approaches}

Clustering-based approaches make the assumption that ``close" data
points tend to have the same label.  Label propagation \cite{Zhu02} is
an algorithm that iteratively adds nearest unlabeled data points to
the set of labeled data.  In a two-label class problem (0 or 1),
initially all unlabeled data points are assigned 0.5 representing
uncertainty in whether that data point belongs to the 0 class or the 1
class.  Until node values converge, the node values are propagated to
their connected nodes representing the unlabeled data points and are
averaged.  Labels are then assigned based on the final value: if the
value is greater than 0.5, then the data point is assigned a 1;
otherwise, it is assigned a 0.

\subsubsection{Self-training}

Self-training uses a model's own predictions as the labels for
retraining \cite{Triguero15}.  A model is initially trained using the
available labeled data.  Unlabeled data is then passed through the
model and assigned the label that the model predicts.  Generally,
labels are only provided to data points where the model has sufficient
confidence.  However, calibrating a model's confidence is not straight
forward \cite{Hyams2017_Arxiv}.  A glaring problem that SUMER attempts
to address is that these methods are not able to correct prediction
errors.  Also, most studies only examine a single iteration and do not
measure the impact of mislabeled data points on subsequent iterations
of self-training.

\subsubsection{Multi-view learning}

Multi-view learning builds on self-training.  Rather than using a
single model to train and label the data, multi-view learning trains
multiple models with different ``views" of the data.  ``Views" of the
data can differ based on the features, data preprocessing, and/or
subsets of the data.  In co-training \cite{Blum1998_COLT}, where there
are two views of the data, data points with confident predictions
according to exactly one of the two models is moved to the training
set for the other model.  In other words, one model provides the
labels for data points about which the other model is uncertain.  This
process is repeated until there are no confident predictions from one
of the classifiers.

There are several variations of multi-view training that build on
co-training.  One of the best known multi-view training methods is
tri-training \cite{Zhou2005_TKDE}, which leverages three
independently-trained models where each initial model is diverse.  For
tri-training, a data point is added to the training set of a model if
the other two models agree on its label.  Like with co-training, this
process is repeated until there are no additional data points added to
a training set.

\subsubsection{Self-ensembling}

Self-ensembling methods are another variation on the multi-view theme
of using model diversity to increase robustness.  The general idea is
to use a single model under different configurations.  There have been
several recent advances in this area focused particularly on deep
learning methods where self-training-like methods are
used \emph{during} the training process.  For example, ladder
networks \cite{Rasmus2015_NIPS} use unlabeled data points with the
goal of making a model more robust to noise.  For each unlabeled
example, noise is added (perturbing the input values) and the example
is assigned the label predicted by the neural network on the clean
version of the example.  Ladder networks are mostly used in computer
vision where many forms of perturbation and data augmentation are
available.

Pseudo-labeling \cite{Lee2013_WREPL} uses self-training in each
training epoch in neural network training.  An initial model is
trained on the labeled set of training examples.  The trained model is
then used to predict the labels of unlabeled training data
(``pseudo-labels"), which is combined with the original labeled data
points and the model is retrained with the pseudo-labeled and the
labeled data.  This is repeated until the model converges.  Temporal
ensembling \cite{Laine2016_Arxiv} builds on pseudo-labels by providing
an exponential moving average of the predictions on the unlabeled data
points as training progresses.  Temporal ensembling also uses a loss
function on the consistency between the network outputs when using
dropout and other regularization techniques.  Mean
teacher \cite{Tarvainen2017_NIPS} is an improvement of temporal
ensembling that stores an exponential moving average of the model
parameters (weights).  There are two conceptual networks: the teacher
network and the student network.  Initially, the teacher network is a
copy of the student network.  Each network will use the same
mini-batch of training data, but the teacher network will add random
augmentation of noise to the inputs.  The (mean) teacher maintains an
exponential moving average of the student network's parameters and
provides a consistency cost between the teacher and student models.
The student network is updated using classification loss.  Currently,
mean teacher provides state-of-the-art results for semi-supervised
learning in the image domain.

\subsubsection{Virtual adversarial training}

The previous approaches used a supervised technique to predict a label
for the unlabeled points.  Virtual Adversarial Training
(VAT) \cite{Miyato2015_ICLR} is an alternative approach that takes
into account the input data distribution irrespective of the class.
The goal of VAT is to make the output distribution of the model smooth
such that the model is not sensitive to small perturbations in the
inputs.  In other words, similar data points should have similar
outputs from the model.  At a high-level, VAT trains a model to make
the outputs of two similar inputs as close as possible.  To do so, VAT
starts with in an input $x$, which it transforms by adding small
perturbations in an adversarial manner (meaning that the the
perturbations encourage large output differences).  With the
adversarial data point, the model weights are updated to minimize the
difference of the output of the original data point and the perturbed
version.  VAT can be used with data sets that are fully-labeled,
partially-labeled, or have no labels.

\subsubsection{Comparison of methods}

\begin{figure}[!h]
	\centering
	\includegraphics[width=.7\textwidth]{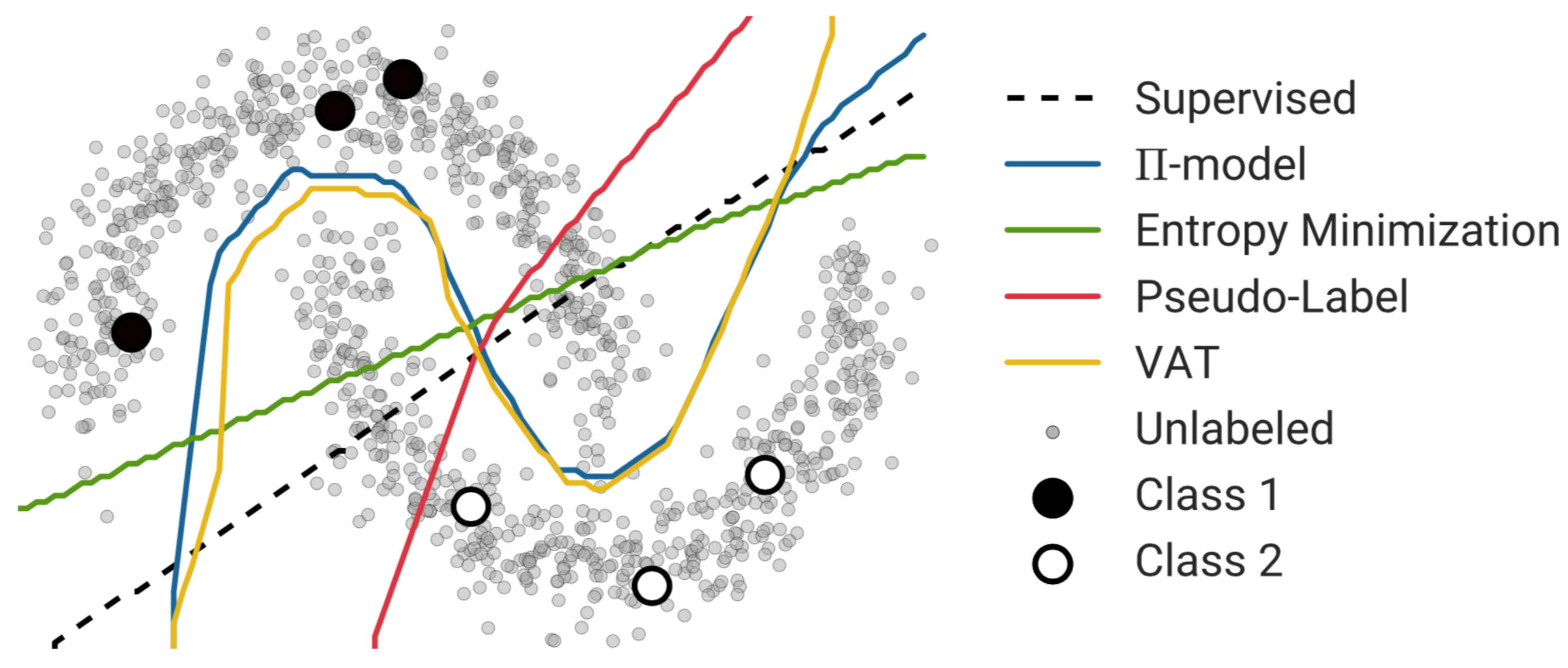} 
	\caption{Comparison of multiple semi-supervised methods on the
          synthetic two-moons data set from Oliver et
          al. \cite{Oliver2018_NIPS}}
	\label{fig:comparison}
\end{figure}

The performance of the techniques is dependent on several factors,
including the initial set of labeled data points.  Recent
work \cite{Oliver2018_NIPS} compared several semi-supervised
techniques to evaluate many real-world scenarios that are not commonly
addressed.  Their findings suggest that temporal ensembling (referred
to as $\pi$-model) and VAT perform the best as illustrated in
Fig.~\ref{fig:comparison}.  As with supervised machine learning
techniques, there are several biases that are present in each of the
techniques that should be considered.  For example, the boundaries
from pseudo-labeling produces more circular clusters of the labeled
data.

\subsection{Label Noise Remediation}
\label{sec:noise}

Learning with noisy labels is a machine learning problem which assumes
that the labeling process (e.g., a human-in-the-loop) that provides
labels for instances will occasionally miscategorize those instances
(i.e., introduce errors). The goal of algorithms used to address label
noise is to induce a model that performs as well as a model that was
built on data with correct labels.

\subsubsection{Estimating noise rates}
\label{sec:related-noise-rates}

For binary classification problems, a machine learning algorithm is
essentially trying to induce a model to separate two distributions,
$P_0$ and $P_1$. When label noise is present, the two distributions
can be viewed as a contaminated mixture of each other \cite{Scott13}:

\begin{align*}
\tilde{P_0} &= (1 - \pi_0) P_0 + \pi_0 P_1 \\ 
\tilde{P_1} &= (1 - \pi_1) P_1 + \pi_1 P_0
\end{align*}

Many label error remediation techniques \cite{Liu16, Northcutt17,
Scott13} attempt to estimate $\pi_0$ and $\pi_1$ directly from the
given contaminated training samples:

\begin{align*}
X^1_0, X^2_0, \ldots, X^{n0}_0 &\sim \tilde{P_0} \\ 
X^1_1, X^2_1, \ldots, X^{n1}_1 &\sim \tilde{P_1} \\ 
\end{align*}

Then, with access to the samples from $\tilde{P_0}$ and $\tilde{P_1}$,
along with the estimated noise rates, $\pi_0$ and $\pi_1$, it should
be possible to recover the true distributions, $P_0$ and $P_1$. Or, at
the very least, it should be possible to induce a machine learning
model as if the true distributions are known. For example, Rank
Pruning \cite{Northcutt17} uses the estimated error rates to prune
(i.e., remove) $\pi_1|\tilde{P_1}|$ and $\pi_0|\tilde{P_0}|$ (where
$|\tilde{P_x}|$ represents the number of instances in class $x$),
which are the least confident data points from the training set, and
then reweights the remaining instances based on those noise rates.


\subsubsection{Dataset augmentation}

Another way to remediate errors in the labels involves altering or
augmenting the training data in intelligent ways (either the instances
or the labels). For example, Kegelmeyer et al.\@ use an ensemble of
anomaly detection models to create new features that have proven
useful in detecting and remediating label noise \cite{Crussell15,
Kegelmeyer15}.  Resampling methods are used to increase diversity and
performance in decision tree
ensembles \cite{Kegelmeyer07}. Synthetically generating data has been
shown to remediate issues such as class skew \cite{Kegelmeyer02}.

Some algorithms augment the labels themselves by attempting to correct
mislabeled instances. For example, Kremer et al.\@ use ideas from
active learning to select instances to relabel, based on those
instances that would have the maximal impact on the
model \cite{Kremer18}.

\subsubsection{Specialized algorithms}

Additionally, specialized algorithms can be derived to address the
issue of label noise directly. For example, label
spreading \cite{Zhou03} relaxes the optimization objective used in
label propagation to allow instances with provided labels to change
labels, which can correct mislabeled instances. Natarajan et al.\@
modify the loss function to address the label noise
issue \cite{Natarajan13}. Their derived loss function also utilizes
noise rates, but assumes that these rates are known in advance. Menon
et al.\@ proved that the balanced error rate and area under the ROC
curve (AUC) are immune to label noise \cite{Menon15}. However,
optimizing these losses might require more complex algorithms or
optimization procedures.

Some algorithms are somewhat more robust to label noise, such as decision trees \cite{Ghosh16} and Random Forests \cite{Frenay14}. The count-based methods used to determine splits and the resampling and randomness used to create the ensembles help to alleviate the effects of label noise. Additionally, there is evidence that deep learning is also quite robust to  non-adversarial label noise \cite{Rolnick17}.

As the main goal of SUMER is to reduce and remediate labeling
errors introduced by the self-updating process, the SUMER framework
involves combining techniques from self-updating / semi-supervised
learning with error remediation techniques, some of which were
described above. Semi-supervised learning and learning with label
noise have been studied for quite a while. However, not all of the
problems associated with these learning tasks have been fully
solved. This section provides some potential research for future
exploration.

%% file: tex/conclusion.tex
In this work, we have experimentally demonstrated the benefit of
self-updating models (SUMs) and self-updating models with error
remediation (SUMER) in both synthetic and realistic environments. In
many cases, SUMs/SUMER may provide improved performance with minimal
downside risk. Our conclusion is that if that environment allows, then
machine learning models should be self-updated and remediated.

This work is the first step toward building a fully autonomous machine
learning (AML) system.  However, we have identified three main
technical problems that must be solved before an end-to-end AML system
can be built:

\begin{enumerate}

\item Identify and characterize changes in label and feature
  distributions in the live stream that the model is making
  predictions on.  Another important aspect would be to identify when
  a new concept or concepts appear that the deployed model has not
  been trained to recognize.  These techniques can be stand-alone or
  integrated into the prediction model \cite{masud2010classification}.

\item Use this change information to suggest the most
  appropriate way to rebuild the model, e.g., incremental or full
  rebuild.  How the model is rebuilt will differ from model-to-model.
  AML is largely model agnostic, but it may be necessary to devise
  algorithms for updating models in the appropriate way if no
  techniques exist for specific models of interest.

\item Augment the retraining dataset to improve the performance of the
  updated model.  For SUMER, we validated the benefits of SUMs and
  label correction techniques.  Future work in this area could look at
  weak supervision, additional label correction techniques (e.g.,
  uncertainty quantification), and using generative models (e.g.,
  Virtual Adversarial Training) to synthesize additional retraining
  data points.  This is a very rich area.  SUMER reduced the risk
  substantially, but there are still lots of potential R\&D
  opportunities.

\end{enumerate}

Future work in AML should also address the challenge of \emph{model
coupling} that appears when the prediction model and the label
correction model provide consistent predictions, i.e., predictions
that are almost always in agreement.  This can be caused by both
models being built using the same view of the data or the models
having the same inductive bias. A view of the data refers to the
feature set created during the feature engineering process. Possible
solutions are to create an independent, auxiliary label correction
model; optimize loss that is immune to noise; and use different views
of the data if possible.  The last potential solution is similar to
the requirement for co-training to have different views of the data.

%% file: tex/acknowledgement.tex
The authors would like to thank our sponsors for supporting this effort and for providing considerable technical,
logistical, and programmatic support over the course of this research.


This paper describes objective technical results and analysis. Any
subjective views or opinions that might be expressed in the paper do
not necessarily represent the views of the U.S. Department of Energy
or the United States Government.

Sandia National Laboratories is a multi-mission laboratory managed and
operated by National Technology and Engineering Solutions of Sandia
LLC, a wholly owned subsidiary of Honeywell International Inc.\ for
the U.S.\ Department of Energy's National Nuclear Security
Administration under contract DE-NA0003525.